\documentclass{seed-survey}

\usepackage{amsmath}
\usepackage{amssymb}
\usepackage{booktabs}
\usepackage{tabularx}
\usepackage{array}
\usepackage{makecell}
\usepackage{pifont}
\usepackage[table]{xcolor}
\usepackage{multirow}
\usepackage{enumitem}
\usepackage{longtable}
\usepackage{float}
\usepackage[T1]{fontenc}
\usepackage{lmodern}
\usepackage{textcomp}
\usepackage{xcolor}
\usepackage{graphicx}
\usepackage{adjustbox}
\usepackage{forest}

\usetikzlibrary{arrows.meta,shapes.geometric,calc}

\providecolor{seedInk}{HTML}{19172F}
\providecolor{seedMuted}{HTML}{6D6A80}
\providecolor{seedPurple}{HTML}{6757D9}
\providecolor{seedDeep}{HTML}{30277A}
\providecolor{seedBlue}{HTML}{477BFF}
\providecolor{seedCyan}{HTML}{35D3CB}
\providecolor{seedCoral}{HTML}{FF6B8A}

\definecolor{wmAmber}{HTML}{E8912F}
\definecolor{wmTeal}{HTML}{16A085}

\colorlet{wmRoot}{seedDeep!78!seedPurple}
\colorlet{wmFound}{seedDeep}      
\colorlet{wmPlau}{seedBlue}       
\colorlet{wmCtrl}{seedCyan}       
\colorlet{wmAct}{seedCoral}       
\colorlet{wmLoop}{seedPurple}     
\colorlet{wmDom}{wmTeal}          
\colorlet{wmData}{wmAmber}        
\colorlet{wmMuted}{seedMuted!65!seedInk}  

\newcommand{\wmfont}{\fontsize{5.3}{6.6}\selectfont}    
\newcommand{\wmfontd}{\fontsize{5.5}{6.9}\selectfont}   
\newcommand{\wmfontb}{\fontsize{5.8}{7.2}\selectfont}   
\newcommand{\wmfontc}{\fontsize{6.8}{8.4}\selectfont}   
\newcommand{\wmfontr}{\fontsize{8.8}{10.8}\selectfont}  

\newcommand{\wmrot}[2]{%
  \rotatebox{90}{\begin{minipage}[c]{#1}\centering #2\end{minipage}}}

\newcommand{\wmreg}{\normalfont\sffamily\mdseries\upshape}
\newcommand{\wmit}{\normalfont\sffamily\mdseries\itshape}

\newcommand{\wmsec}[1]{{\wmit\S#1}}

\forestset{
  wmtree/.style={
    for tree={
      grow'=east,
      parent anchor=east,
      child anchor=west,
      anchor=west,
      draw,
      rounded corners=1.7pt,
      line width=0.5pt,
      font=\sffamily\wmfont,
      inner xsep=3.2pt,
      inner ysep=2.2pt,
      edge={line width=0.5pt},
      edge path'={ (!u.parent anchor) -- ++(3pt,0) |- (.child anchor) },
      s sep=2.2pt,
      l sep=6pt,
    },
  },
  wmroot/.style={
    fill=wmRoot, draw=wmRoot, text=white,
    font=\sffamily\bfseries\wmfontr,
    align=center,
    inner xsep=4.5pt, inner ysep=5pt,
    rounded corners=3pt,
  },
  wmlvlone/.style n args={1}{
    fill=#1!52!seedInk, draw=#1!52!seedInk, text=white,
    font=\sffamily\bfseries\wmfontc,
    align=center,
    inner xsep=3.2pt, inner ysep=4pt,
    rounded corners=2.4pt,
    edge={#1!52!seedInk, line width=0.8pt},
    for descendants={edge={#1!45!seedInk, line width=0.5pt}},
  },
  wmlvltwo/.style n args={2}{
    fill=#1!24, draw=#1!60!seedInk, text=seedInk,
    font=\sffamily\bfseries\wmfontb,
    text width=#2,
    text ragged,
    inner xsep=3.4pt, inner ysep=3.2pt,
    rounded corners=2.2pt,
  },
  wmlvlthree/.style n args={2}{
    fill=#1!12, draw=#1!55!seedInk!70!white, text=seedInk,
    font=\sffamily\bfseries\wmfontd,
    text width=#2,
    text ragged,
    inner xsep=3.2pt, inner ysep=2.6pt,
  },
  wmleaf/.style n args={2}{
    fill=#1!5, draw=#1!40!seedInk!50!white, text=seedInk,
    font=\sffamily\wmfont,
    text width=#2,
    text ragged,
    inner xsep=3.2pt, inner ysep=2.2pt,
  },
}

\newcommand{\Plevel}{\textsc{Plausible}}
\newcommand{\Clevel}{\textsc{Controllable}}
\newcommand{\Alevel}{\textsc{Actionable}}

\title{World Models for Embodied Intelligence: From Plausible to Controllable to Actionable}

\authors{
Nanjie Yao\textsuperscript{1} \quad 
Hao Wang\textsuperscript{1*} \quad 
Chong Cheng\textsuperscript{1} \quad 
Zhikang Chen\textsuperscript{2} \quad 
Wenzhe Li\textsuperscript{3} \\
Jiafei Lyu\textsuperscript{4} \quad 
Li Shen\textsuperscript{5} \quad 
Peilin Zhao\textsuperscript{6} \quad 
Zongqing Lu\textsuperscript{7} \quad 
Gao Huang\textsuperscript{8} \quad 
Steven Hoi\textsuperscript{9,10} \quad 
Dacheng Tao\textsuperscript{11} \quad 
Deheng Ye\textsuperscript{11*}
}

\affiliations{
$^{1}$HKUST(GZ)\quad
$^{2}$University of Oxford\quad
$^{3}$Princeton University\quad
$^{4}$Tencent\quad
$^{5}$Sun Yat-sen University\quad
$^{6}$Shanghai Jiao Tong University\quad
$^{7}$Peking University\quad
$^{8}$Tsinghua University\quad
$^{9}$Alibaba Group\quad
$^{10}$Singapore Management University\quad
$^{11}$Nanyang Technological University\quad

}
\authornote{*: Corresponding Author}

\surveyabstract{%
World models connect perception and decision-making in embodied intelligence by maintaining hidden state, anticipating consequences, comparing interventions, and adapting when execution departs from expectations. Although progress is often measured by visual fidelity, their value lies in improving behavior. Before reaching for a cup, a person anticipates its weight and resistance to grasping, shaping the hand before contact. Such anticipation is coarse and rarely pictorial, yet it guides action. This raises a central question: which predictive capabilities improve behavior? Existing surveys, organized by architecture, output modality, or application domain, leave this question implicit. We introduce three progressively stronger capability levels: \Plevel{} models preserve task-relevant temporal, geometric, or physical structure; \Clevel{} models additionally predict how interventions alter that structure; and \Alevel{} models translate predictions into measurable gains in planning, action, learning, evaluation, verification, recovery, or data selection. We complement this hierarchy with a 3 × 4 matrix crossing geometry, physics, and action grounding with improvement loops centered on data, rewards, policies, and the model itself. Using this framework, we survey manipulation, navigation, locomotion, autonomous driving, and general embodied learning, tracing technical progressions, clarifying capability requirements, and examining datasets, benchmarks, and evaluation protocols. We identify challenges in long-horizon consistency, uncertainty calibration, causal intervention testing, latency, verification and recovery, and cross-embodiment transfer. This perspective shifts evaluation from visual plausibility toward whether predictions capture task-relevant state, reflect intervention effects, and improve the closed-loop behavior of embodied agents.

}

\project{\href{https://3dagentworld.github.io/EmbodiedWM/}{\texttt{Project Page}}}
\date{\today}

\affiliationlogos{%

  \includegraphics[height=0.70cm]{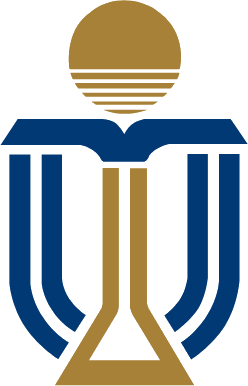}\hspace{0.18cm}%
  \includegraphics[height=0.70cm]{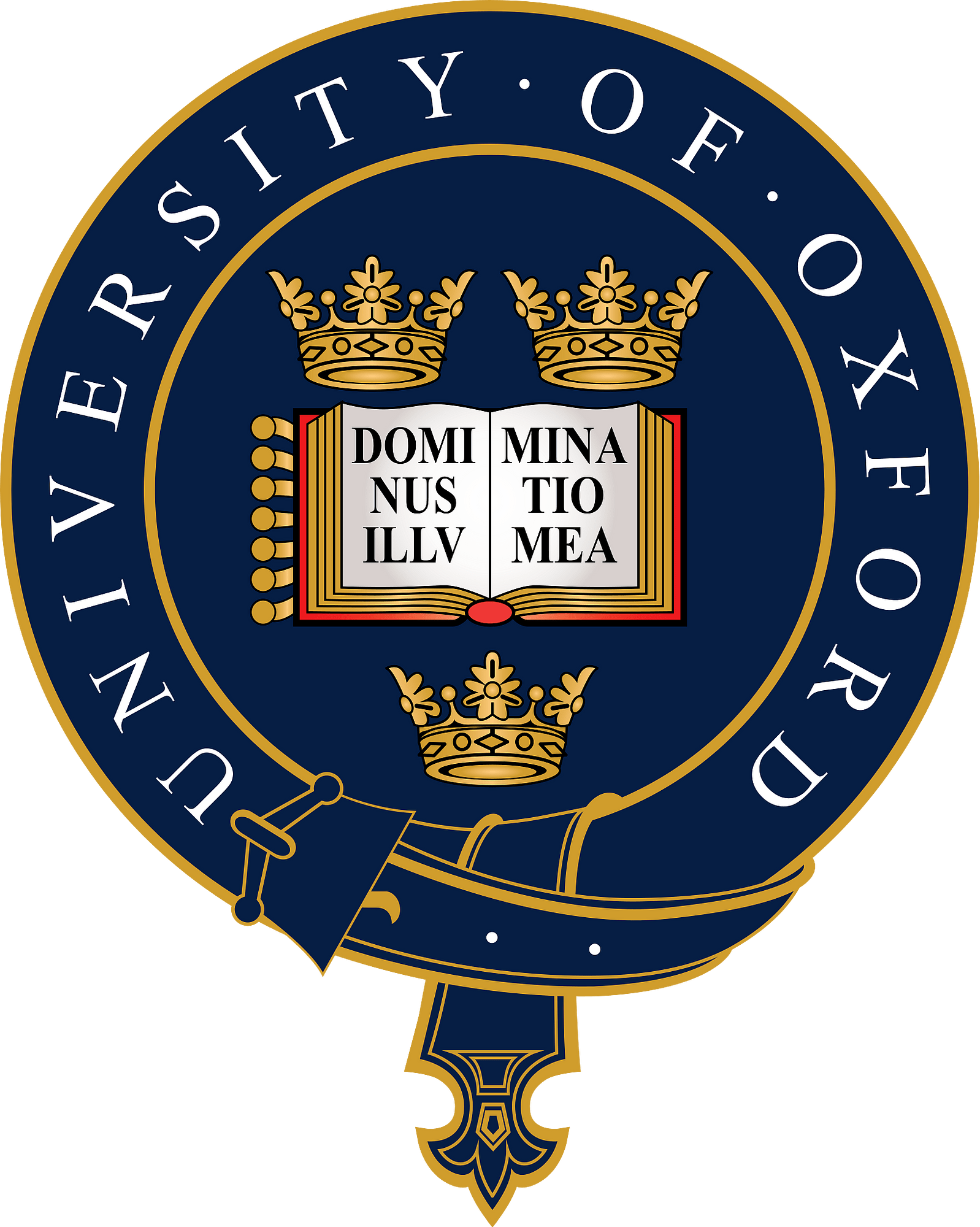}\hspace{0.18cm}%
  \includegraphics[height=0.70cm]{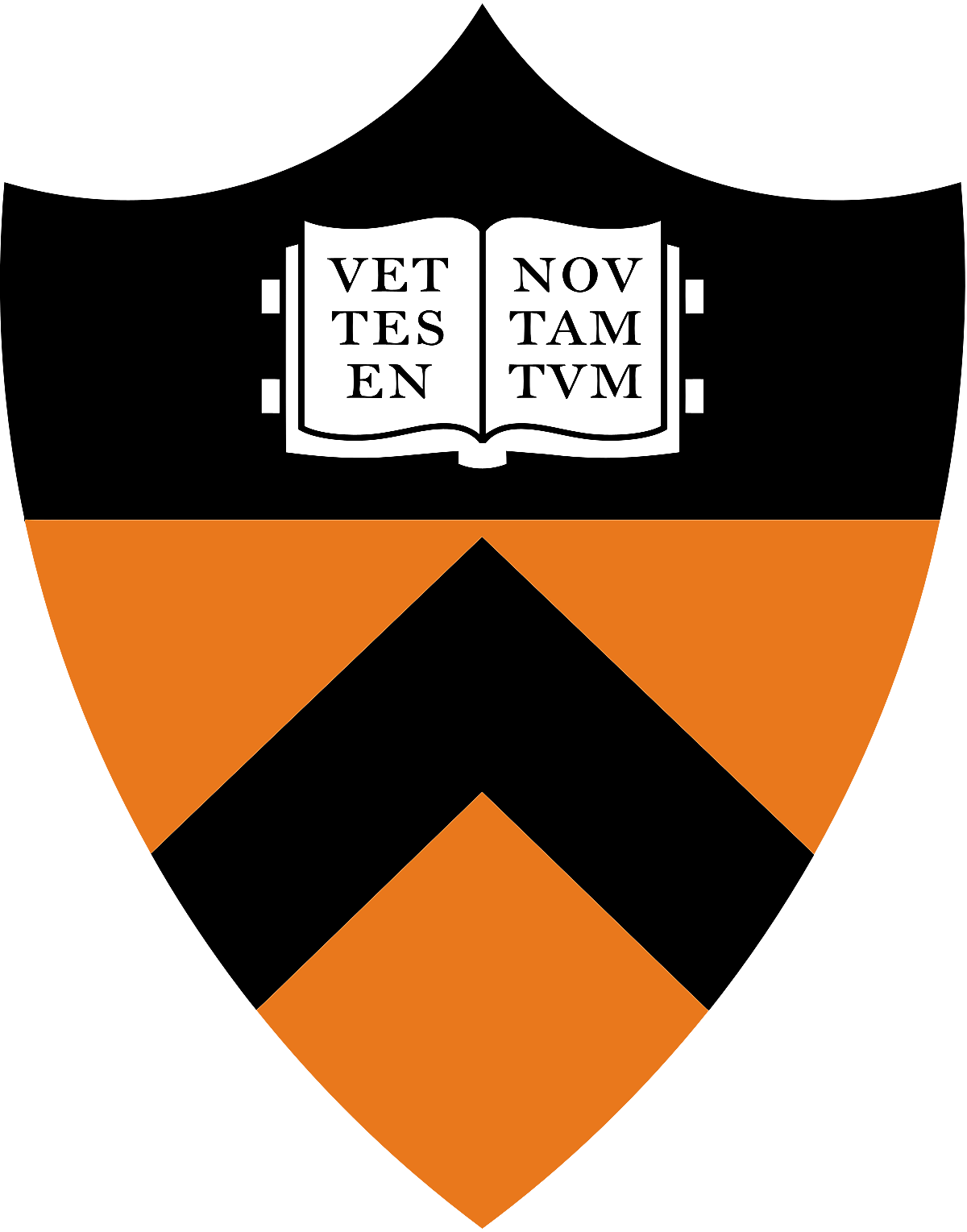}\hspace{0.18cm}%
  \includegraphics[height=0.70cm]{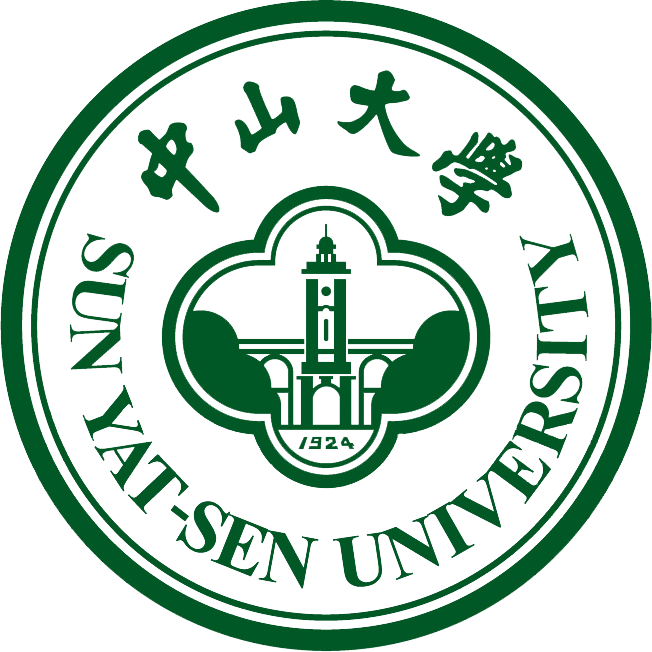}\hspace{0.18cm}%
  \includegraphics[height=0.70cm]{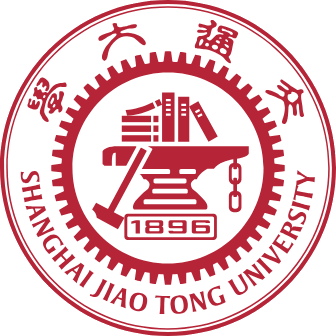}\hspace{0.18cm}%
  \includegraphics[height=0.70cm]{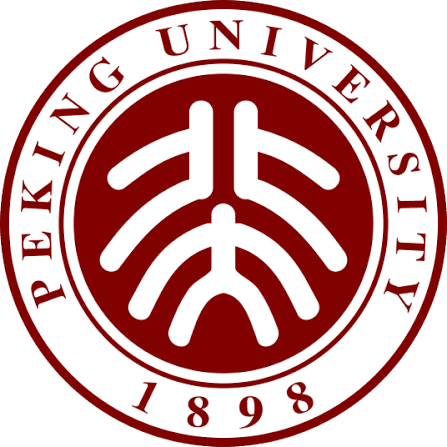}\hspace{0.18cm}%
  \includegraphics[height=0.70cm]{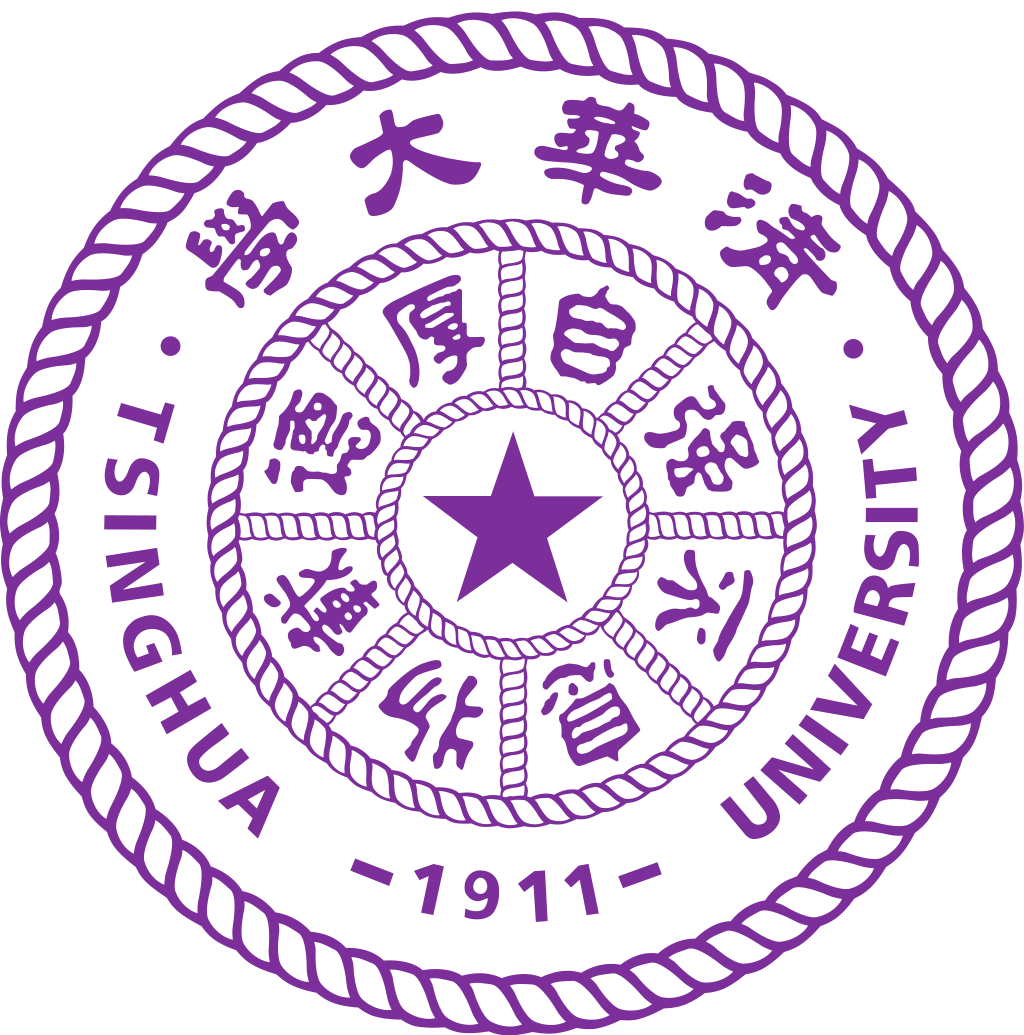}\hspace{0.18cm}%
  \includegraphics[height=0.70cm]{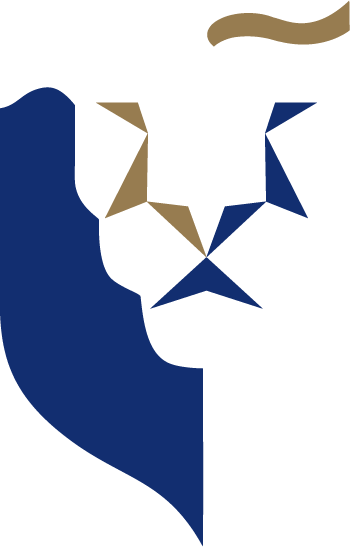}\hspace{0.18cm}%
  \includegraphics[height=0.70cm]{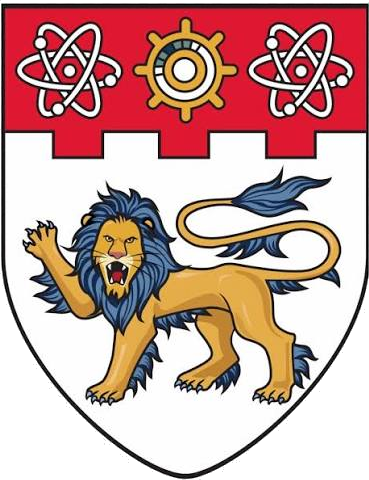}\hspace{0.18cm}%
  \includegraphics[height=0.70cm]{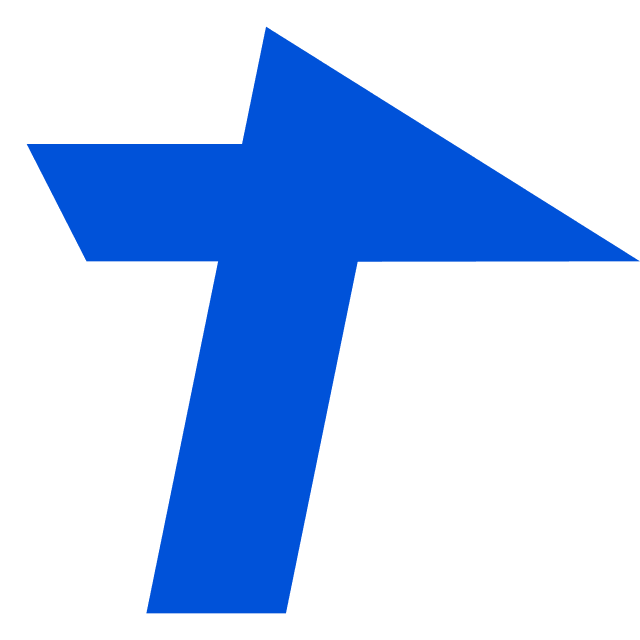}\hspace{0.18cm}%
  \includegraphics[height=0.70cm]{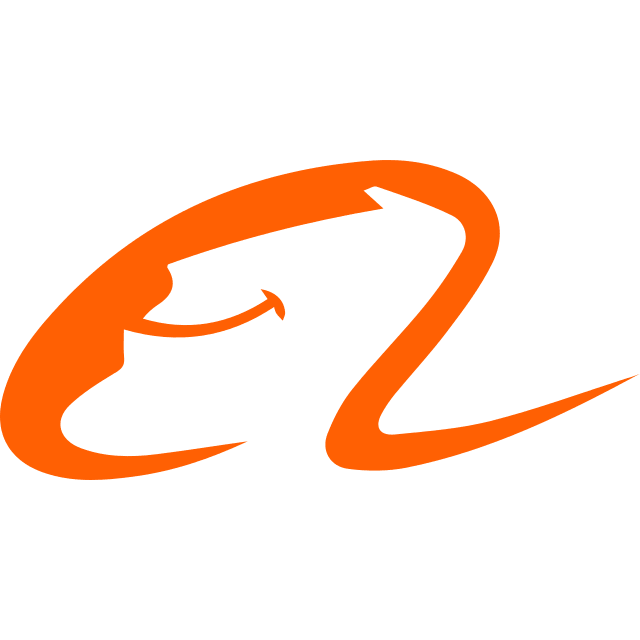}\hspace{0.18cm}%
}

\begin{document}

\maketitle

\makeseedtoc

%

\section{Introduction}
\label{sec:introduction}

Research in embodied intelligence increasingly explores generalist policies that couple visual perception, language understanding, and action generation. Vision--language--action (VLA) models exemplify this trend: they map multimodal observations and instructions to robot actions, with the goal of broad task coverage and flexible instruction following~\cite{Shao2025arxiv3073,Zhong2025arxiv1925}.
However, a direct observation-to-action mapping provides no explicit predictive mechanism for maintaining task-relevant state, anticipating delayed consequences, or comparing alternative interventions. These capabilities matter most in partially observed~\cite{Hafner2018arxiv4551,Huang2025arxiv7868,Huang2025arxiv0041}, contact-rich~\cite{He2026arxiv8555,Zheng2026arxiv9201,Li2025arxiv6693}, and long-horizon environments~\cite{Ma2026arxiv2295,Liu2026arxiv6789,Guo2024arxiv1522}, where an agent may need to reason about unobserved state, predict the consequences of candidate actions, and revise its behavior when execution departs from expectation.

\begin{figure}
    \centering
    \includegraphics[width=1.0\linewidth]{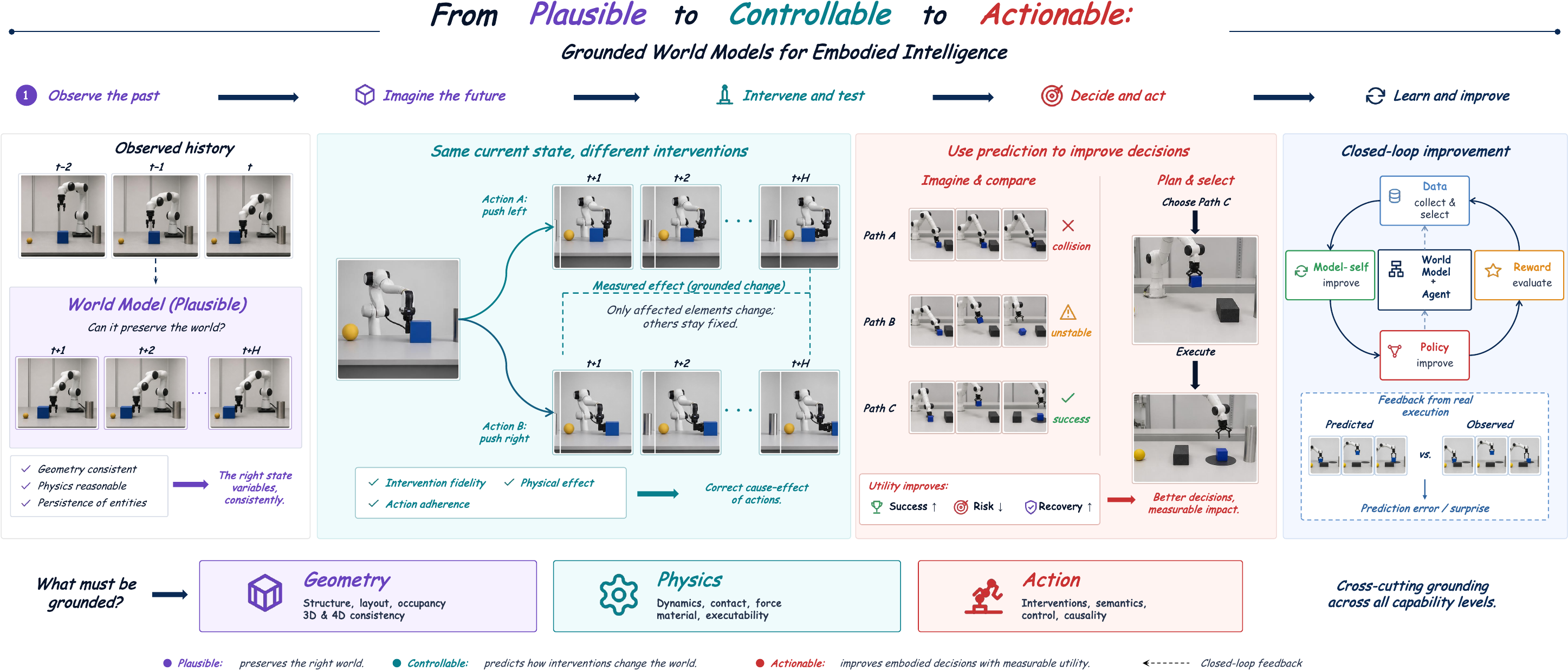}
    \caption{\textbf{Overview of this survey paper.} The framework is illustrated through a robotic-manipulation example. Starting from observed history, Plausible models preserve task-relevant states. Controllable models predict how interventions change the environment, while Actionable models use these predictions to improve decisions with measurable utility. Geometry, physics, and action ground all three levels. Feedback from real execution supports the closed-loop improvement of data, rewards, policies, and the world model itself.}
    \label{fig:overview}
\end{figure}

These requirements motivate \emph{world models}, which represent how an environment evolves and mediate between perception and decision making. For embodied intelligence, modeling the world is a means to better behavior: its value lies in helping an agent anticipate consequences, compare interventions, and make better decisions. This perspective shifts the central question from how realistically a model reconstructs the world to what its predictions enable an embodied agent to do. Figure~\ref{fig:overview} summarizes this progression from predictive representation to grounded intervention and closed-loop decision making. Useful world modeling therefore entails three progressively stronger requirements. A model must first preserve the \emph{right state}, maintaining task-relevant temporal, geometric, or physical structure over time; it must then respond to interventions in the \emph{right way}, so that different actions induce corresponding changes in predicted futures; and its predictions must finally be \emph{useful for behavior}, producing measurable improvements in downstream decision making, learning, evaluation, verification, or recovery.

World modeling for embodied agents has developed along several connected directions. Latent-dynamics and value-equivalent models learn compact predictive states for search, planning, and policy optimization~\cite{Ha2018arxiv0122,Hafner2019arxiv1603,Schrittwieser2019arxiv8265}. Visual and structured predictors broaden predictive representations to visual features, occupancy, and persistent 3D/4D scene state~\cite{Bardes2024arxiv8471,Zheng2023arxiv6038,Zhou2025arxiv5495}. Physics-aware models further introduce interaction variables and constraints such as motion, contact, force, and material response~\cite{Yang2024arxiv8410,Li2025arxiv6693,He2026arxiv8555}. More recently, action-conditioned and joint world action models have coupled prediction more closely with control by conditioning future observations on actions or generating futures and actions together~\cite{Guo2024arxiv8179,Zhu2025arxiv2792,Cheang2024arxiv6158}. Decision-facing systems then place these predictors inside planning~\cite{Khorrambakht2025arxiv3077}, policy learning~\cite{Zheng2025arxiv5659}, functional evaluation~\cite{Li2025arxiv9017}, and execution-time verification and correction~\cite{Liu2026arxiv6789,Pan2026arxiv1804}. Together, these lines span a conceptual progression from passive future generation to closed-loop uses of prediction in embodied interaction.

\begingroup
\small
\begin{longtable}{@{}>{\raggedright\arraybackslash}p{0.27\textwidth}>{\raggedright\arraybackslash}p{0.25\textwidth}>{\raggedright\arraybackslash}p{0.40\textwidth}@{}}
\caption{Closest survey anchors and the scope added by this review.}\label{tab:survey-comparison}\\
\toprule
\textbf{Survey} & \textbf{Primary view} & \textbf{Perspective emphasized here} \\
\midrule
\endfirsthead
\toprule
\textbf{Survey} & \textbf{Primary view} & \textbf{Perspective emphasize here} \\
\midrule
\endhead
Initial Driving WM Survey~\cite{Guan2024arxiv2622} & Perception, memory, prediction, and control in driving & Adds cross-domain intervention and decision gains. \\
Video Generation and Driving WMs~\cite{Fu2024arxiv2914} & Relation between generative video and driving simulation & Adds contact, recovery, and non-video predictive states. \\
Understanding or Predicting?~\cite{Ding2024arxiv4499} & World understanding versus future prediction & Treats acting as a measured capability rather than an application label. \\
Driving WM Survey~\cite{Feng2025arxiv1260} & Scene generation and behavior planning & Extends action and physics beyond vehicle trajectories. \\
Simulators and WMs~\cite{Long2025arxiv0917} & Physical simulators, learned simulators, and robot intelligence levels & Adds a fine-grained action interface and capability ladder. \\
VLA Action Tokenization~\cite{Zhong2025arxiv1925} & Language, code, trajectory, latent, and raw-action tokens & Adds future stability, physical grounding, and prediction feedback. \\
VLM-based VLA Survey~\cite{Shao2025arxiv3073} & Monolithic and hierarchical VLA systems & Makes the world model a tested capability rather than one optional module. \\
Embodied AI WM Survey~\cite{Li2025arxiv6732} & Functional, temporal, and spatial representations & Connects reward, recovery, and model update within one framework. \\
Video Generation in Robotics~\cite{Mei2026arxiv7823} & Robot applications of video generation & Adds latent, 3D/4D, force, and decision-centered capability. \\
WM for Robot Learning~\cite{Hou2026arxiv0080} & WM--policy coupling and learned simulators & Adds three capability levels and four improvement loops. \\
WM for Manipulation~\cite{Wang2026arxiv0113} & Video, latent, motion, 3D/4D, and physics models & Extends beyond manipulation and proposes capability-aligned evaluation across domains. \\
WAM Frontier~\cite{Wang2026arxiv2090} & Cascaded and joint WAM architectures & Adds post-action verification, recovery, and model-self update. \\
WAM Survey~\cite{Shen2026arxiv0781} & Substrate, backbone, action coupling, and deployment & Places WAMs inside a broader plausible--controllable--actionable ladder. \\
\bottomrule
\end{longtable}
\endgroup

This progression, however, is not the primary organizing principle of most existing surveys. Related surveys instead adopt distinct axes, including architecture, representation, application, and embodiment. General world-model surveys~\cite{Ding2024arxiv4499,Li2025arxiv6732} emphasize predictive representation and world understanding. Reviews of model-based reinforcement learning, robot learning, and learned simulation~\cite{Moerland2020arxiv6712,Hou2026arxiv0080,Long2025arxiv0917} focus on dynamics learning, planning, policy optimization, and simulator construction. Autonomous-driving surveys~\cite{Guan2024arxiv2622,Fu2024arxiv2914,Feng2025arxiv1260} organize work around scene generation, behavior prediction, and planning, whereas robotics-oriented video-generation reviews~\cite{Mei2026arxiv7823} emphasize data synthesis, action prediction, and reinforcement learning. VLA surveys~\cite{Shao2025arxiv3073,Zhong2025arxiv1925} primarily classify policy architectures and action representations, while surveys of manipulation and world action models~\cite{Wang2026arxiv0113,Wang2026arxiv2090,Shen2026arxiv0781} study predictive targets, action interfaces, model--policy coupling, and deployment regimes. These perspectives are complementary, but their organizing axes do not directly capture the capability progression from preserving predictive state, to modeling intervention effects, to demonstrating downstream decision utility.

To make this capability progression explicit, we organize the literature around a decision-centered ladder: \Plevel{} $\rightarrow$ \Clevel{} $\rightarrow$ \Alevel{}. At the \Plevel{} level, a world model preserves task-relevant temporal, geometric, or physical structure. At the \Clevel{} level, it additionally predicts how interventions change that structure. At the \Alevel{} level, its predictions are used to improve a downstream decision or update, yielding measurable benefits. The levels describe progressively stronger capabilities rather than mutually exclusive architecture classes, allowing methods with different architectures, representations, and embodiments to be examined within a common framework. Table~\ref{tab:survey-comparison} summarizes how this organizing perspective complements closely related surveys and scope papers.

We complement this ladder with two cross-cutting views. The first identifies \emph{what grounds the prediction}: the geometry, physics, and action dimensions specify the represented structure, feasible dynamics, and intervention effects on which a capability claim depends. The second identifies \emph{how prediction improves the system}: data, reward, policy, and model-self loops describe where predictive knowledge enters learning and interaction. Their intersections yield a unified $3\times4$ grounding--improvement matrix that connects representational content to closed-loop use. We apply this framework to research spanning manipulation, navigation, locomotion, autonomous driving, and general embodied learning.

In summary, this survey makes four contributions.

\begin{itemize}[leftmargin=*]
    \item We formulate \Plevel{}, \Clevel{}, and \Alevel{} world modeling as a decision-centered capability ladder spanning state consistency, intervention fidelity, and decision utility, providing a unified framework for organizing methods across architectures, representations, and application domains.

    
    \item We connect the capability ladder to a $3\times4$ matrix that crosses the geometry, physics, and action grounding dimensions with data, reward, policy, and model-self improvement loops, thereby linking predictive content to the mechanism through which it affects an embodied agent.

    \item We organize representative methods into coherent technical progressions around the three capability levels and analyze their cross-cutting connections across grounding dimensions, improvement loops, and embodiments, covering manipulation, navigation, locomotion, and autonomous driving.

    \item We review datasets and benchmarks, outline evaluation protocols aligned with the three capability levels, and identify open problems in state consistency, intervention fidelity, and decision utility, with particular emphasis on uncertainty, latency, verification, recovery, and cross-embodiment transfer.
    
    
\end{itemize}

\paragraph{Roadmap.} Section~\ref{sec:background} establishes the foundations and scope of world modeling for embodied learning. Section~\ref{sec:taxonomy} formalizes the capability ladder and the grounding--improvement matrix. Sections~\ref{sec:plausible}--\ref{sec:actionable} review the progression from state-consistent prediction to intervention-aware modeling and decision-facing use. Section~\ref{sec:loops} examines four feedback mechanisms through which world-model predictions shape data acquisition, reward and critic signals, policies, and updates to the models themselves. Section~\ref{sec:domains} compares embodiment-specific requirements, Section~\ref{sec:data-evaluation} reviews datasets and evaluation protocols, and Section~\ref{sec:open-problems} develops the research challenges implied by the framework.

\section{Foundations, Definitions, and Boundaries}
\label{sec:background}

\subsection{From Learned Dynamics to World Models}

Modern world models are rooted in three closely related ideas: learning environment dynamics, predicting future states under possible actions, and improving behavior through imagined experience. Early learned dynamics models mainly focused on approximating state transitions for planning and control. However, embodied agents require more than accurate transition prediction.
They must preserve task-relevant information, capture the effects of interventions, and provide predictions that improve downstream decisions. A world model should therefore be defined by whether it represents the variables that matter for interaction rather than by whether it reconstructs observations.

\paragraph{From visual prediction to latent dynamics.}
Early work on visual foresight explored whether future observations could be predicted from robot actions. Deep Visual Foresight~\cite{Finn2016arxiv0696} predicts future pixels conditioned on robot actions and uses the predictions inside model-predictive control. Visual Foresight~\cite{Ebert2018arxiv0568} extends this idea with self-supervised robot data and repeated replanning. These models provide an explicit connection between prediction and control: actions modify the predicted future, and the predicted future guides subsequent actions. PETS~\cite{Chua2018arxiv2114} introduces probabilistic ensembles and trajectory sampling, making uncertainty an essential component of planning. PlaNet~\cite{Hafner2018arxiv4551} further shifts prediction from pixels to compact latent states, allowing planning without decoding each predicted
latent state into an observation.

\paragraph{Latent imagination and predictive representations.}
The development of latent world models demonstrates that useful prediction can be performed in abstract state spaces. World Models~\cite{Ha2018arxiv0122} introduces a perception--memory--controller framework and trains controllers inside learned latent environments. Dreamer~\cite{Hafner2019arxiv1603} learns policies from imagined trajectories generated by latent dynamics, while DreamerV2~\cite{Hafner2020arxiv2193} and DreamerV3~\cite{Hafner2023arxiv4104} improve scalability through discrete representations and robust training procedures. Later methods broaden this line along complementary directions. Dreaming~\cite{Okada2020arxiv4535} removes observation reconstruction and learns decision-relevant latent dynamics through a decoder-free contrastive objective, while TransDreamer~\cite{Chen2022arxiv9481}, IRIS~\cite{Micheli2022arxiv0588}, and TWM~\cite{Robine2023arxiv7109}, further explore transformer-based and token-based predictive models. Together, these studies establish an important principle: a world model does not need to generate realistic observations if its internal state preserves the information required for prediction and decision making.

\paragraph{From predictive models to decision-oriented models.}
Another line of research shows that world models can be useful even without explicit observation prediction. The Predictron~\cite{Silver2016arxiv8810} learns internal rewards, discounts, and values for planning. Value Prediction Networks~\cite{Oh2017arxiv3497} learn abstract transitions that preserve value-related structure. MuZero~\cite{Schrittwieser2019arxiv8265} combines learned dynamics with tree search without reconstructing observations, and EfficientZero~\cite{Ye2021arxiv0210} improves low-data learning through consistency and value-prefix objectives. These methods shift the goal of world modeling from reproducing the external world to preserving the internal structure required for effective decisions.

\paragraph{Controlling model exploitation and uncertainty.}
A learned model can improve decision making only when prediction errors are properly managed. SimPLe~\cite{Kaiser2019arxiv0374} trains policies with synthetic rollouts generated by a video model, while MBPO~\cite{Janner2019arxiv8253} controls model bias by limiting rollout horizons. MOPO~\cite{Yu2020arxiv3239} and COMBO~\cite{Yu2021arxiv8363} incorporate conservative objectives for offline reinforcement learning. Iso-Dream~\cite{Pan2022arxiv3817} explicitly separates controllable and non-controllable dynamics. TD-MPC~\cite{Hansen2022arxiv4955} and TD-MPC2~\cite{Hansen2023arxiv6828} combine latent dynamics with efficient planning across tasks and embodiments, while DayDreamer~\cite{Wu2022arxiv4176} demonstrates latent imagination on physical robots. These works highlight several practical requirements for embodied world models: uncertainty estimation, short-horizon prediction, value-aware optimization, and continuous grounding with real experience.

\paragraph{Towards controllable and actionable world models.}
Recent research increasingly connects future prediction with action generation. UniPi~\cite{Du2023arxiv0111} generates task-related videos and converts them into actions through inverse dynamics, providing an explicit video-to-action interface. Subsequent approaches introduce richer action conditioning, latent action representations, joint video--action generation, and direct action decoding. These developments indicate a transition from passive future prediction toward interactive world models that can represent how actions change the environment and how such predictions influence behavior. We therefore analyze these methods according to their capability for controllability and actionability rather than treating video generation as an independent model category.

\subsection{World Models: Definition and Scope}

In this survey, we use the term \emph{world model} in an embodied and decision-oriented sense. A world model is a learned or hybrid predictive model of agent--environment dynamics: given an interaction
history and, optionally, a candidate sequence of actions, it predicts how task-relevant aspects of the environment may evolve. Its defining property is therefore the prediction of temporal change in a form that can support embodied reasoning or decision making, rather than the reconstruction of a complete external world.

Formally, at time $t$, an agent receives an observation $o_t$, executes an action $a_t$, and may receive a task signal $r_t$. We summarize the available interaction context as
\[
h_t = (o_{1:t}, a_{1:t-1}, r_{1:t-1}, g),
\]
where $g$ denotes an optional language instruction, goal state, or task specification. An action-conditioned world model may predict
\[
p_\theta\!\left(
z_{t+1:t+H}, r_{t:t+H-1}
\mid h_t, a_{t:t+H-1}
\right),
\]
where $H$ is the prediction horizon and $z$ denotes the modeled state. This formulation is intentionally agnostic to the state representation: $z$ may correspond to pixels, visual tokens, learned features, latent states, objects, geometry, physical variables, symbolic states, or decision-equivalent quantities such as rewards and values. The history may also be compressed into a belief state $s_t=f_\phi(h_t)$, from which future states and task signals are predicted. Decoding the state back  into observations is optional.

We use this definition functionally rather than architecturally. Predictions may be consumed by a planner, used to train a policy through imagined trajectories, queried as a simulator, or coupled directly with action generation. These uses do not constitute mutually exclusive model classes. For example, a learned simulator is a world model exposed through a repeatable rollout interface, whereas a \emph{world action model} tightly couples future prediction and action generation through a cascade, shared representation, or joint generative process. Similarly, reward, value, and critic models fall within our scope only when they evaluate predicted transitions, imagined trajectories, or model uncertainty rather than only the current observation.

This functional definition also clarifies the boundary with neighboring models. A video generation model is an observation-space world-model candidate when its generated future represents the temporal evolution of an embodied task environment. However, visual plausibility alone does not demonstrate that the model captures the effects of robot actions or benefits downstream decisions. Conversely, a vision-language-action model is primarily a policy model. A purely reactive VLA without an identifiable predictive mechanism is outside our operational scope; it enters the taxonomy when a predicted future state, transition, or outcome influences action generation, policy learning, candidate selection, or execution.

Accordingly, we distinguish the existence of a predictive model from the strength of the capability supporting it. Prediction of a \emph{plausible} future establishes neither faithful action control nor decision utility. We use \emph{controllable} for models whose predictions respond correctly to robot-relevant interventions, and \emph{actionable} for models whose predictions produce measurable improvements in decision making or learning. Section~\ref{sec:taxonomy} formalizes these distinctions through the Plausible--Controllable--Actionable capability ladder.

\subsection{Predictive Representations}

World models can be described along several independent axes. The first axis is the space in which future states are predicted. Observation-space models predict pixels or visual tokens, which makes their outputs easy to inspect but can require substantial computation~\cite{Finn2016arxiv0696,Ebert2018arxiv0568}. Latent-space models instead predict compact continuous or discrete features that retain information useful for dynamics and control~\cite{Hafner2018arxiv4551,Hafner2020arxiv2193,Micheli2022arxiv0588}. Structured models organize the predicted state around objects, relations, maps, or symbolic variables. These choices define the form of the representation, but not its capability.

The second axis is the semantic content of the predicted state. Appearance-based models predict how a scene may look. Geometry-grounded models predict depth, pose, point clouds, occupancy, scene flow, or persistent 3D/4D structure~\cite{Zheng2023arxiv6038,Zhou2025arxiv5495,cheng2026horizonstream,11092901,zhang2025sat,shen2025smpl,shu2026fastanimate,cheng2026longstreamlongsequencestreamingautoregressive,hu2025vggt4dminingmotioncues}. Physics-grounded models predict motion, contact, force, deformation, material response, or physical feasibility~\cite{Yang2024arxiv8410,Li2025arxiv6693,He2026arxiv8555}. Task-grounded models may predict rewards, values, termination, progress, or risk~\cite{Silver2016arxiv8810,Oh2017arxiv3497,Schrittwieser2019arxiv8265}. A model can combine several targets through shared dynamics and separate prediction heads.

The third axis describes the temporal and action interface. A model may predict one transition, an action chunk, a subgoal, or a long trajectory. It may be deterministic or represent uncertainty through distributions, ensembles, or samples. Its input may include no action, explicit controls, trajectories, latent actions, or high-level skills. Some models predict only future states, whereas world action models couple future prediction with action generation~\cite{Du2023arxiv0111,Zhu2025arxiv2792}. These properties determine how a model can be queried and used. They do not by themselves establish prediction quality or decision value.

Representation should therefore be evaluated separately from capability. A pixel model can support planning or verification, while a 3D model can remain inaccurate or unused by the agent. Larger backbones and longer rollouts also do not imply stronger world modeling. In our framework, plausibility requires stable prediction of task-relevant structure. Controllability requires correct responses to changes in action. Actionability requires measurable gains in planning, learning, evaluation, verification, recovery, or data selection.

\begin{takeaway}
World models vary in representation space, semantic content, temporal horizon, uncertainty modeling, and action interface. These choices shape what a model can predict and how its predictions can be used, but they do not determine its capability. The three capability levels
correspond to stable task-relevant prediction for plausibility, correct responses to action changes for controllability, and measurable downstream gains for actionability.
\end{takeaway}

\begin{figure*}[p]
  \begin{adjustbox}{max width=\textwidth,max totalheight=0.84\textheight}
  \begin{forest}
    wmtree,
[\wmrot{4.2cm}{\textsc{Grounded World Models}\\ {\wmreg\wmfontb for Embodied Intelligence}}, wmroot, calign=child, calign child=2
  [\wmrot{4.0cm}{\textsc{Plausible}\\ {\wmreg\wmfontb State consistency}\\ \wmsec{4}}, wmlvlone={wmPlau}, tier=capability
    [{Latent and\\ Object State \wmsec{4.1}}, wmlvltwo={wmPlau}{1.6cm}
      [{Object slots}, wmlvlthree={wmPlau}{1.5cm}
        [{C-SWM~\cite{Kipf2019arxiv2247}, SlotFormer~\cite{Wu2022arxiv5861}, Hindsight for Foresight~\cite{Nematollahi2020hindsight}}, wmleaf={wmPlau}{9.7cm}]]
      [{Joint\\ embeddings}, wmlvlthree={wmPlau}{1.5cm}
        [{I-JEPA~\cite{Assran2023ijepa}, V-JEPA~\cite{Bardes2024arxiv8471}, V-JEPA~2.1~\cite{MurLabadia2026arxiv4482}, LeWorldModel~\cite{Maes2026arxiv9312}}, wmleaf={wmPlau}{9.7cm}]]
    ]
    [{Metric Scene\\ State \wmsec{4.2}}, wmlvltwo={wmPlau}{1.6cm}
      [{Occupancy\\ and BEV}, wmlvlthree={wmPlau}{1.5cm}
        [{OccWorld~\cite{Zheng2023arxiv6038}, MUVO~\cite{Bogdoll2023arxiv1762}, DriveWorld~\cite{Min2024arxiv4390}, BEVWorld~\cite{Zhang2024arxiv5679}, DynamicCity~\cite{Bian2024arxiv8084}, OccSora~\cite{Wang2024arxiv0337}, Occupancy WM for Robots~\cite{Zhang2025arxiv5512}}, wmleaf={wmPlau}{9.7cm}]]
      [{Points and\\ rendering}, wmlvlthree={wmPlau}{1.5cm}
        [{ViDAR~\cite{Yang2023arxiv7655}, RenderWorld~\cite{Yan2024arxiv1356}, GaussianWorld~\cite{Zuo2024arxiv0373}, Geometry Forcing~\cite{Wu2025arxiv7982}, Geometry-aware 4D Video~\cite{Liu2025arxiv1099}, Aether~\cite{Team2025arxiv8945}, TesserAct~\cite{Zhen2025arxiv0995}, PointWorld~\cite{Huang2026arxiv3782}}, wmleaf={wmPlau}{9.7cm}]]
    ]
    [{Physical\\ Dynamics \wmsec{4.3}}, wmlvltwo={wmPlau}{1.6cm}
      [{Benchmarks}, wmlvlthree={wmPlau}{1.5cm}
        [{IntPhys~\cite{Riochet2018arxiv7616}, PHYRE~\cite{Bakhtin2019arxiv5656}, CLEVRER~\cite{Yi2019arxiv1442}, Physion~\cite{Bear2021arxiv8261}}, wmleaf={wmPlau}{9.7cm}]]
      [{Structured\\ predictors}, wmlvlthree={wmPlau}{1.5cm}
        [{PhyDNet~\cite{Guen2020arxiv1460}, OP3~\cite{Veerapaneni2019arxiv2827}, DrivePhysica~\cite{Yang2024arxiv8410}, FOLIAGE~\cite{Liu2025arxiv3173}, ParticleFormer~\cite{Huang2025arxiv3126}, RoboScape~\cite{Shang2025arxiv3135}, PhysWorld~\cite{Mao2025arxiv7416}, WoW~\cite{Chi2025arxiv2642}, ABot-PhysWorld~\cite{Chen2026arxiv3376}, Cosmos WFM~\cite{NVIDIA2025arxiv3575}}, wmleaf={wmPlau}{9.7cm}]]
    ]
    [{Persistent\\ Memory \wmsec{4.4}}, wmlvltwo={wmPlau}{1.6cm}
      [{Persistent Embodied WM~\cite{Zhou2025arxiv5495}, Long-term Spatial Memory~\cite{Wu2025arxiv5284}, InfinityDrive~\cite{Guo2024arxiv1522}, Long-Context State-Space Video WM~\cite{Po2025arxiv0171}, LongDWM~\cite{Wang2025arxiv1546}, WorldDreamer~\cite{Wang2024arxiv9985}}, wmleaf={wmPlau}{11.65cm}]
    ]
  ]
  [\wmrot{4.8cm}{\textsc{Controllable}\\ {\wmreg\wmfontb Intervention fidelity}\\ \wmsec{5}}, wmlvlone={wmCtrl}, tier=capability
    [{Action\\ Conditioning \wmsec{5.1}}, wmlvltwo={wmCtrl}{1.6cm}
      [{Manipulation\\ and navigation}, wmlvlthree={wmCtrl}{1.5cm}
        [{SAVP~\cite{Lee2018arxiv1523}, FitVid~\cite{Babaeizadeh2021arxiv3195}, PACT~\cite{Bonatti2022arxiv1133}, IRASim~\cite{Zhu2025irasim}, Navigation World Models~\cite{Bar2025nwm}}, wmleaf={wmCtrl}{9.7cm}]]
      [{Driving}, wmlvlthree={wmCtrl}{1.5cm}
        [{GAIA-1~\cite{Hu2023arxiv7080}, DriveDreamer~\cite{Wang2023arxiv9777}, Vista~\cite{Gao2024arxiv7398}, Copilot4D~\cite{Zhang2023arxiv1017}, ACT-Bench~\cite{Arai2024arxiv5337}}, wmleaf={wmCtrl}{9.7cm}]]
    ]
    [{Open and Latent\\ Control \wmsec{5.2}}, wmlvltwo={wmCtrl}{1.6cm}
      [{Latent\\ actions}, wmlvlthree={wmCtrl}{1.5cm}
        [{Genie~\cite{Bruce2024arxiv5391}, LAPA~\cite{Ye2024arxiv1758}, AdaWorld~\cite{Gao2025arxiv8938}, LDA-1B~\cite{Lyu2026arxiv2215}}, wmleaf={wmCtrl}{9.7cm}]]
      [{Open worlds\\ and language}, wmlvlthree={wmCtrl}{1.5cm}
        [{UniSim~\cite{Yang2023arxiv6114}, MineWorld~\cite{Guo2025arxiv8388}, Yume~\cite{Mao2025arxiv7744}, Pandora~\cite{Xiang2024arxiv9455}, RoboDreamer~\cite{Zhou2024arxiv2377}, ManipDreamer~\cite{Li2025arxiv6464}, Object-Centric WM~\cite{Jeong2025arxiv6170}, Goal-VLA~\cite{Chen2025arxiv3919}}, wmleaf={wmCtrl}{9.7cm}]]
    ]
    [{Structure-\\ Grounded \wmsec{5.3}}, wmlvltwo={wmCtrl}{1.6cm}
      [{Geometry and\\ trajectory}, wmlvlthree={wmCtrl}{1.5cm}
        [{MagicDrive~\cite{Gao2023arxiv2601}, DOME~\cite{Gu2024arxiv0429}, Mask2IV~\cite{Li2025arxiv3135b}, RoboMaster~\cite{Fu2026robomaster}, 3DFlowAction~\cite{Zhi2025arxiv6199}, GAF~\cite{Chai2025arxiv4135}, 3D-VLA~\cite{Zhen2024arxiv9631}}, wmleaf={wmCtrl}{9.7cm}]]
      [{Contact and\\ force}, wmlvlthree={wmCtrl}{1.5cm}
        [{DreMa~\cite{Barcellona2024arxiv4957}, DexSim2Real$^2$~\cite{Jiang2024arxiv8750}, PIN-WM~\cite{Li2025arxiv6693}, Physically Embodied Gaussian Splatting~\cite{AbouChakra2024arxiv0788}, OmniVTA~\cite{Zheng2026arxiv9201}}, wmleaf={wmCtrl}{9.7cm}]]
    ]
    [{Joint\\ World--Action \wmsec{5.4}}, wmlvltwo={wmCtrl}{1.6cm}
      [{Interactive\\ rollout}, wmlvlthree={wmCtrl}{1.5cm}
        [{Ctrl-World~\cite{Guo2025arxiv0125}, iVideoGPT~\cite{Wu2024arxiv5223}, Vid2World~\cite{Huang2025arxiv4357}, DreamDojo~\cite{Gao2026arxiv6949}, Mem-World~\cite{Zheng2026arxiv8960}, EnerVerse~\cite{Huang2025arxiv1895}, EnerVerse-AC~\cite{Jiang2025arxiv9723}}, wmleaf={wmCtrl}{9.7cm}]]
      [{Shared\\ backbones}, wmlvlthree={wmCtrl}{1.5cm}
        [{GR-1~\cite{Wu2023arxiv3139}, GR-2~\cite{Cheang2024arxiv6158}, Video Prediction Policy~\cite{Hu2024arxiv4803}, PAD~\cite{Guo2024arxiv8179}, Unified Video Action~\cite{Li2025arxiv0200}, Unified World Models~\cite{Zhu2025arxiv2792}, WorldVLA~\cite{Cen2025arxiv1539}, DyWA~\cite{Lyu2025arxiv6806}, EVA~\cite{Wang2026arxiv7808}}, wmleaf={wmCtrl}{9.7cm}]]
    ]
  ]
  [\wmrot{6.6cm}{\textsc{Actionable}\\ {\wmreg\wmfontb Decision utility}\\ \wmsec{6}}, wmlvlone={wmAct}, tier=capability
    [{Planning and\\ Selection \wmsec{6.1}}, wmlvltwo={wmAct}{1.6cm}
      [{Search and\\ MPC}, wmlvlthree={wmAct}{1.5cm}
        [{Video Language Planning~\cite{Du2023arxiv0625}, Large Video Planner~\cite{Chen2025arxiv5840}, Safe MPC~\cite{Koller2019arxiv2189}, Drive-WM~\cite{Wang2023arxiv7918}, WorldPlanner~\cite{Khorrambakht2025arxiv3077}, STORM~\cite{Lin2025arxiv8477}, GHIL-Glue~\cite{Hatch2025ghilglue}, Inference-Time Enhancement~\cite{Qi2026ralitps}, Reimagination~\cite{Chen2025arxiv6565}}, wmleaf={wmAct}{9.7cm}]]
      [{Value and\\ structured plans}, wmlvlthree={wmAct}{1.5cm}
        [{VIPER~\cite{Escontrela2023arxiv4343}, DINO-WM~\cite{Zhou2024arxiv4983}, From Pixels to Predicates~\cite{Athalye2026ral}, Exopredicator~\cite{Liang2026exopredicator}, StructVLA~\cite{Jin2026arxiv2553}, World Action Planner~\cite{Zhang2026arxiv7599}, World-Value-Action~\cite{Li2026arxiv4732}, EvolvingAgent~\cite{Feng2025arxiv5907}}, wmleaf={wmAct}{9.7cm}]]
    ]
    [{Policy and \\ Action Decoding \wmsec{6.2}}, wmlvltwo={wmAct}{1.6cm}
      [{Action from\\ imagination}, wmlvlthree={wmAct}{1.5cm}
        [{ContextWM~\cite{Wu2023arxiv8499}, Dreamitate~\cite{Liang2024arxiv6862}, NovaFlow~\cite{Li2026novaflow}, TC-IDM~\cite{Mi2026arxiv8323}, Video Generators are Robot Policies~\cite{Liang2025arxiv0795}, Grounding Video Models to Actions~\cite{Luo2025grounding}, VidMan~\cite{Wen2024arxiv9153}}, wmleaf={wmAct}{9.7cm}]]
      [{Predictive\\ policies}, wmlvlthree={wmAct}{1.5cm}
        [{FLARE~\cite{Zheng2025arxiv5659}, JEPA-VLA~\cite{Miao2026arxiv1832}, FRAPPE~\cite{Zhao2026arxiv7259}, UP-VLA~\cite{Zhang2025arxiv8867}, X-MOBILITY~\cite{Liu2024arxiv7491}, DiT4DiT~\cite{Ma2026arxiv0448}, DreamZero~\cite{Ye2026arxiv5922}, WSA$_1$~\cite{Jiang2026arxiv3941}, DriveWorld-VLA~\cite{jia2026arxiv6521}, LaWAM~\cite{Chen2026arxiv5768}, ForeWAM~\cite{Huang2026arxiv1605}, AHEAD~\cite{Syed2026arxiv2486}, AHA-WAM~\cite{Cai2026arxiv9811}, Fast-WAM~\cite{Yuan2026arxiv6666}, $\tau_0$-WM~\cite{Zhou2026arxiv1027}}, wmleaf={wmAct}{9.7cm}]]
    ]
    [{Learned\\ Simulators \wmsec{6.3}}, wmlvltwo={wmAct}{1.6cm}
      [{Policy\\ training}, wmlvlthree={wmAct}{1.5cm}
        [{SafeDreamer~\cite{Huang2023arxiv7176}, Think2Drive~\cite{Li2024arxiv6720}, VLM-SAFE~\cite{Qu2025arxiv6377}, DriveArena~\cite{Yang2024arxiv0415}, Robotic World Model~\cite{Li2025arxiv0100}, LUMOS~\cite{Nematollahi2025arxiv0370}, SENSEI~\cite{Sancaktar2025arxiv1584}, EmbodieDreamer~\cite{Wang2025arxiv5198}, ReDRAW~\cite{Lanier2025arxiv2252}, World-Env~\cite{Xiao2025arxiv4948}, WMPO~\cite{Zhu2026wmpo}, VLA-RFT~\cite{Li2025arxiv0406}, DiWA~\cite{Chandra2025diwa}, Prophesying~\cite{Zhang2025arxiv0633}, VLAW~\cite{Guo2026arxiv2063}}, wmleaf={wmAct}{9.7cm}]]
      [{Policy\\ evaluation}, wmlvlthree={wmAct}{1.5cm}
        [{SIMPLER~\cite{Li2024arxiv5941}, WorldEval~\cite{Li2025arxiv9017}, WorldGym~\cite{Quevedo2025arxiv0613}, RoboWorld~\cite{Jeon2026arxiv1060}, Genie Envisioner~\cite{Liao2025arxiv5635}, Veo world simulator~\cite{GeminiRobotics2025arxiv0675}, WorldSimBench~\cite{Qin2024arxiv8072}, EWMBench~\cite{Yue2025arxiv9694}, WorldArena~\cite{Shang2026arxiv8971}, RoboWM-Bench~\cite{Jiang2026arxiv9092}, MiraBench~\cite{Yang2026arxiv9360}}, wmleaf={wmAct}{9.7cm}]]
    ]
    [{Verification and\\ Recovery \wmsec{6.4}}, wmlvltwo={wmAct}{1.6cm}
      [{Diagnosis}, wmlvlthree={wmAct}{1.5cm}
        [{RoboFAC~\cite{Ye2025arxiv2224}, ViFailback~\cite{Zeng2025arxiv2787}, PhysicalAgent~\cite{Lykov2025arxiv3903}, ST-WAM~\cite{Wang2026arxiv8993}}, wmleaf={wmAct}{9.7cm}]]
      [{Runtime\\ verification}, wmlvlthree={wmAct}{1.5cm}
        [{CheckVLA~\cite{Liu2026arxiv6789}, Foresight~\cite{Zhang2026arxiv3085}, FoMo-FD~\cite{Huang2026arxiv7511}, Bimanual Failure Detection~\cite{Ward2026arxiv6987}, WM Failure Classification~\cite{Ho2026arxiv6182}, ActFovea~\cite{Yu2026arxiv9169}, FAWAM~\cite{He2026arxiv8555}}, wmleaf={wmAct}{9.7cm}]]
      [{Correction and\\ recovery}, wmlvlthree={wmAct}{1.5cm}
        [{When to Trust Imagination~\cite{Wang2026arxiv6222}, TempoWAM~\cite{Ye2026arxiv9492}, VLA-Corrector~\cite{Pan2026arxiv1804}, AdaReP~\cite{Cheng2026arxiv3079}, Feedback World Model~\cite{An2026arxiv5705}, FutureRTC~\cite{Jiang2026arxiv4008}, WAMs in Real Time~\cite{Team2026arxiv1880}, PLanAR~\cite{Guo2026arxiv1662}, See, Plan, Rewind~\cite{Dai2026arxiv9292}, CycleVLA~\cite{Ma2026arxiv2295}, B2FF~\cite{Shin2026arxiv9258}}, wmleaf={wmAct}{9.7cm}]]
    ]
  ]
]
  \end{forest}
  \end{adjustbox}
  \caption{%
   \textbf{Taxonomy of grounded world models for embodied intelligence.}
    Each coloured spine is one rung of the capability ladder, the first tinted column its subsections, the second a thematic group, and the pale cards representative works of that group. Each reference in this figure appears once under a primary narrative role; placement does not certify that all capability requirements have been tested. The rungs are cumulative: \textcolor{seedBlue!55!seedInk}{\textsc{Plausible}} requires task-relevant state consistency over the tested horizon and conditions; \textcolor{seedCyan!55!seedInk}{\textsc{Controllable}} adds correct grounded effects of paired interventions; \textcolor{seedCoral!45!seedInk}{\textsc{Actionable}} adds a measured decision gain under matched budgets. }
  \label{fig:taxonomy}
\end{figure*}
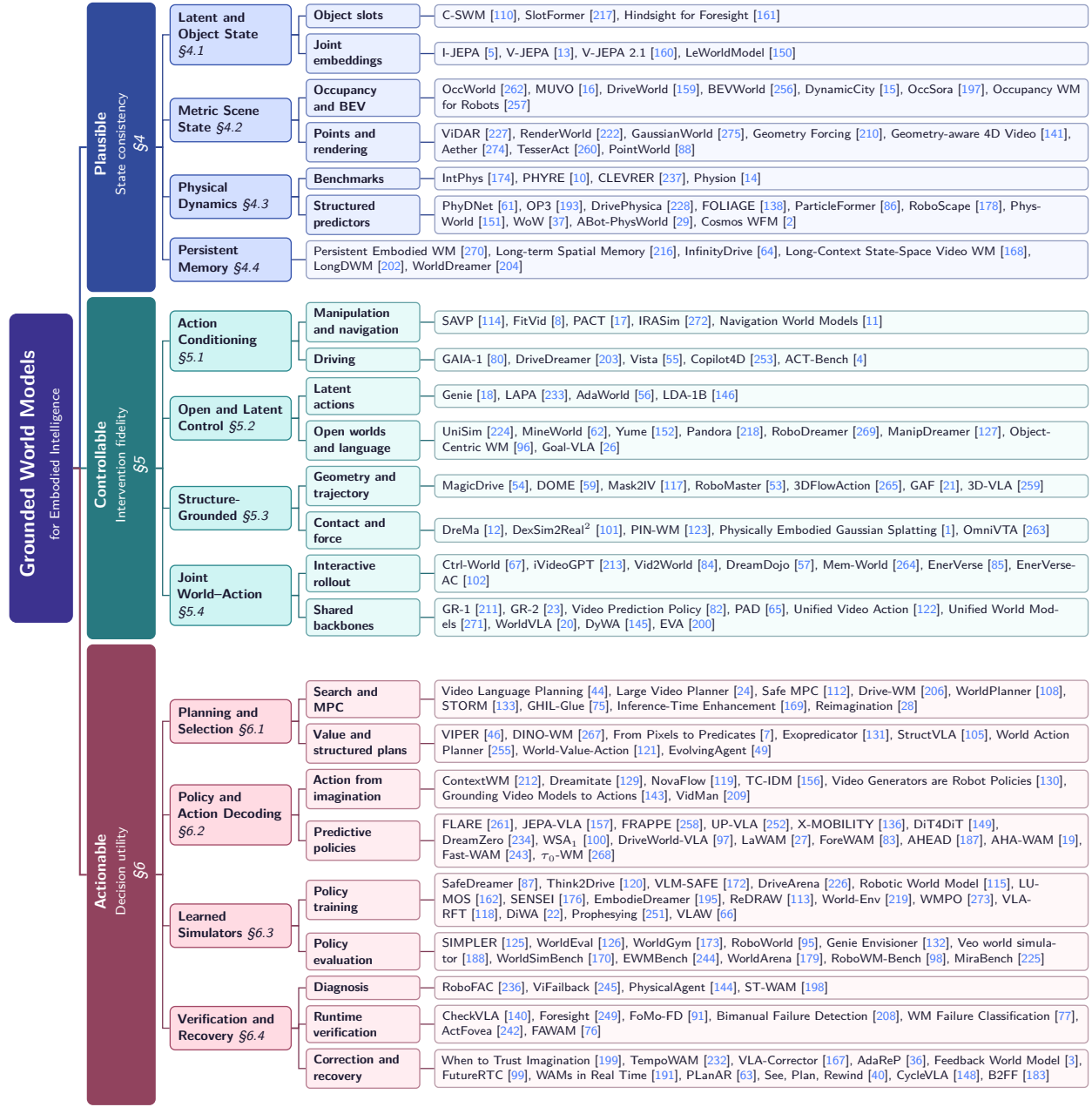

\section{Unified Taxonomy}
\label{sec:taxonomy}

Our taxonomy separates the strength of a world-model claim from the design choices used to realize it. The cumulative \Plevel{}--\Clevel{}--\Alevel{} ladder distinguishes state consistency, intervention fidelity, and decision utility. Two cross-cutting views then explain which state variables a model predicts and how those predictions affect the agent. Geometry, physics, and action grounding identify the state variables that carry the claim, while data, reward, policy, and model-self loops identify the update path through which prediction changes the system. Because these views are multi-label descriptors, they complement rather than subdivide the three capability levels.

The framework is guided by six research questions.

\begin{enumerate}[leftmargin=*,label=\textbf{RQ\arabic*:}]
    \item Which capabilities distinguish plausible, controllable, and actionable models, and which capabilities remain untested?
    \item How do geometry, physics, and action grounding change the state variables and tests used at each capability level?
    \item How do prediction target, horizon, memory, uncertainty, coupling, and deployment affect prediction accuracy and downstream use at each capability level?
    \item How do data, reward, policy, and model-self loops use grounded predictions, and when does a loop yield measured improvement?
    \item Which grounded capabilities transfer across manipulation, navigation, locomotion, and autonomous driving?
    \item Which missing links block the transition from \Plevel{} to \Clevel{} and from \Clevel{} to \Alevel{}?
\end{enumerate}

Sections~\ref{sec:plausible}--\ref{sec:actionable} address the first three questions through capability definitions and method comparisons. Section~\ref{sec:loops} addresses RQ4, Section~\ref{sec:domains} addresses RQ5, and Section~\ref{sec:open-problems} addresses RQ6.

\subsection{A Grounding-Conditioned Capability Ladder}

The Plausible–Controllable–Actionable ladder specifies the strength of a world-model claim, but a capability claim is meaningful only relative to the structure being predicted and the intervention under which it is tested. A model may generate visually plausible futures while failing to preserve spatial relations, obey physical interaction constraints, or respond correctly to a commanded action. We therefore use grounding to denote an explicit and testable correspondence between a model’s predictions and measurable properties of embodied interaction.

We focus on three grounding requirements because they capture three fundamental questions of embodied prediction. Geometry-grounded evaluation asks whether the model preserves what exists and where it is, including identity, pose, depth, occlusion, viewpoint consistency, scene flow, and persistent 3D/4D structure. Physics grounding asks whether the predicted state can evolve through feasible interactions, including motion, collision, contact, friction, force, compliance, and material response. Action grounding asks whether an agent’s intervention is correctly identified and linked to its consequences, including action identity, timing, magnitude, composition, coordinate frame, and embodiment. Geometry therefore specifies the structure of the predicted state, physics constrains its possible evolution, and action grounding links interventions to state transitions.


\paragraph{Plausible.}
A plausible model predicts a future that is consistent with the observed history and the relevant structure of the environment. The claim is stronger than visual realism. Geometry-grounded evaluation tests identity, pose, occupancy, occlusion, and multi-view consistency. Physics-grounded evaluation tests motion, collision, contact, support, and material response. Action-grounded evaluation can test the continuity of agent state and observed consequences, but it does not yet establish counterfactual control. The minimum setting is an open-loop or offline test that measures at least one task-relevant state property across time.

Let $g\in\{\mathrm{geo},\mathrm{phys},\mathrm{act}\}$ index the grounding, let $C_g(z)$ denote grounding-specific constraints, and let $d_g$ measure error in the corresponding state. A useful plausibility test reports
\begin{equation}
    \mathbb{E}[d_g(\hat{z}_{t+h},z_{t+h})],
    \quad
    \Pr[C_g(\hat{z}_{t+1:t+H})=1],
    \quad h=1,\ldots,H,
    \label{eq:plausibility}
\end{equation}
as a function of horizon, scene shift, or uncertainty. Reporting the curve rather than one aggregate score makes drift visible.

\paragraph{Controllable.}
A controllable model represents how an intervention changes the future. Geometry-grounded control tests requested changes in camera, ego, object, or trajectory state. Physics-grounded control tests contact, force, compliance, or material response under intervention. Action-grounded control tests whether identity, timing, magnitude, composition, and embodiment produce the correct effect. For two action sequences $a$ and $a'$, the model should predict distinct grounded effects when the true effects are distinct:
\begin{equation}
    p_\theta(z'_g\mid h_t,\operatorname{do}(a))
    \neq
    p_\theta(z'_g\mid h_t,\operatorname{do}(a')).
    \label{eq:controllability}
\end{equation}
Here, $\operatorname{do}(a)$ denotes a commanded intervention. Conditioning on recorded actions does not by itself isolate action effects, because unobserved conditions may influence both action selection and the outcome.
The predicted contrast must match held-out environment transitions. Changing a prompt or producing a different-looking sample is not sufficient.

\paragraph{Actionable.}
An actionable model changes a downstream decision or update and improves a measured outcome. Geometry becomes actionable when it improves reachability, collision avoidance, mapping, or path selection. Physics becomes actionable when it improves contact, force control, safety, or executability. Action grounding becomes actionable when it improves planning, ranking, policy learning, verification, recovery, or data acquisition. Let $\mathcal{D}$ be a planner, policy, critic, evaluator, verifier, or data selector. The model is actionable when
\begin{equation}
    \mathcal{D}(h_t,p_\theta)\neq \mathcal{D}(h_t,\varnothing)
    \quad\text{and}\quad
    \mathbb{E}[U(\mathcal{D}(h_t,p_\theta))]
    >
    \mathbb{E}[U(\mathcal{D}(h_t,\varnothing))],
    \label{eq:actionability}
\end{equation}
under a matched controller and compute budget. The utility $U$ may be task return, success, policy-ranking quality, safety, recovery, real-data efficiency, or persistent improvement. A qualitative imagined rollout does not meet this condition.

\begin{seedbox}{Why the ladder is cumulative}
Plausibility tests whether task-relevant predictions remain consistent. Controllability tests whether candidate actions produce the expected grounded effects. Actionability tests whether those branches improve an outcome under the task objective and system budget. A method may demonstrate a higher-level use while leaving a lower-level capability untested; the untested state properties or action responses should be reported explicitly rather than inferred from downstream success.
\end{seedbox}

\subsection{The Grounding--Improvement Matrix}
\label{sec:grounding-improvement-matrix}

The capability ladder specifies what a world model demonstrates, whereas the grounding--improvement matrix identifies where its predictions enter the learning and decision cycle. The rows of the matrix indicate the grounded content of prediction, namely geometry, physics, or action effects. The columns indicate the system component that the prediction informs or changes. The \emph{data loop} changes the experience available for learning through data acquisition, generation, or curation. The \emph{reward loop} changes rewards, values, critics, or verification signals. The \emph{policy loop} changes action selection or policy parameters through planning, imitation learning, or reinforcement learning. The \emph{model-self loop} updates the predictor, its persistent state, or its memory using measured prediction error and deployment feedback.

Table~\ref{tab:grounding-improvement-matrix} crosses these two views. Each cell describes a possible interface between a grounded prediction and an improvement path. The cells are neither capability levels nor mutually exclusive paper categories: a method may involve several grounding dimensions and participate in multiple loops.

\begin{table}[t]
    \centering
    \caption{%
        Grounding--improvement matrix. Rows identify the grounded content of
        prediction, and columns identify the component of the learning or
        decision cycle that the prediction informs or changes. Each cell
        represents a possible interaction rather than a capability level or
        a mutually exclusive method category.
    }
    \label{tab:grounding-improvement-matrix}
    \small
    \setlength{\tabcolsep}{4pt}
    \renewcommand{\arraystretch}{1.18}
    \begin{tabularx}{\textwidth}{
        @{}
        >{\raggedright\arraybackslash}p{0.10\textwidth}
        >{\raggedright\arraybackslash}X
        >{\raggedright\arraybackslash}X
        >{\raggedright\arraybackslash}X
        >{\raggedright\arraybackslash}X
        @{}
    }
        \toprule
        \textbf{Grounding}
        & \textbf{Data loop}
        & \textbf{Reward loop}
        & \textbf{Policy loop}
        & \textbf{Model-self loop} \\
        \midrule

        \textbf{Geometry}
        & Acquire viewpoints, occlusions, and topology changes that reduce
          spatial ambiguity.
        & Score collision, visibility, reachability, and spatial progress.
        & Select or learn actions that preserve feasible paths and object
          relations.
        & Correct maps, identities, and persistent scene state after observed
          mismatch. \\

        \textbf{Physics}
        & Acquire contact, slip, deformation, force, and rare-failure
          transitions.
        & Score stability, physical feasibility, energy, and force limits.
        & Select or learn actions that avoid unsafe or physically unsupported
          outcomes.
        & Correct local dynamics and contact behavior while testing for
          regression. \\

        \textbf{Action}
        & Acquire paired interventions that vary command, timing, magnitude,
          composition, or embodiment.
        & Score command adherence, task progress, value, and risk.
        & Select or learn actions from supported counterfactual outcomes.
        & Correct action semantics, delay, and action-conditioned transition
          bias. \\

        \bottomrule
    \end{tabularx}
\end{table}

The four paths are often coupled rather than independent. Data selected or generated by a world model may train a critic, which changes the policy and therefore the state distribution subsequently used to update the world model. The matrix records the interfaces through which such interactions occur; Section~\ref{sec:loops} examines their behavior over repeated updates.

Occupying a matrix cell does not by itself establish actionability. A grounded prediction may inform one of these paths even when only state consistency or action-response accuracy has been tested. It becomes actionable when its use produces a measured downstream improvement over an appropriate matched baseline.

\begin{figure}
    \centering
    \includegraphics[width=1.0\linewidth]{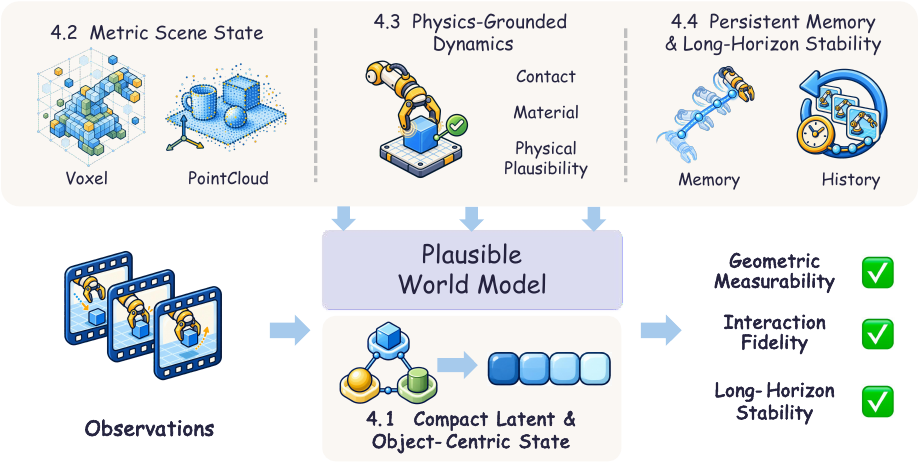}
    \caption{\textbf{Plausible World Model.} Starting from observations, prediction moves from compact latent or object-centric states (Section~\ref{sec:plausible-latent}) to metric scene representations such as voxel grids and point clouds (Section~\ref{sec:plausible-metric}), then to dynamics constrained by contact and material response (Section ~\ref{sec:plausible-physics}), and finally to persistent memory and long-horizon state consistency (Section~\ref{sec:plausible-memory}). These additions progressively make geometry, interaction feasibility, and persistence testable. The progression is conceptual rather than chronological, and the illustrated representations are examples rather than components required in every model.}
    \label{fig:plausible-overview}
\end{figure}

\section{Plausible World Models}
\label{sec:plausible}

A \Plevel{} world model preserves the structure that a later intervention will act on. This requirement is wider than image quality. It covers object identity, spatial layout, occlusion, physical regularity, memory, and stability across a rollout. General video generators such as CogVideoX~\cite{Yang2024cogvideox} show that long and detailed futures can be synthesized at scale. Sharpness alone does not tell an evaluator whether an object kept its identity or whether a wall stayed in place. This section therefore asks what the predicted state makes measurable, and whether that measurement survives time, occlusion, and distribution shift.

We organize the literature into four stages, as shown in Fig.~\ref{fig:plausible-overview}. They are ordered by what each step added to the predicted state, not by publication year, because the stages overlap in time. The first stage moves the predictive target off pixels and onto a compact latent or object-centric state. The second writes the future into a metric frame, so that error can be measured in physical units. The third constrains how the state is allowed to evolve. The fourth requires the same state to still be present at the end of a long rollout. 


\subsection{Compact Latent and Object-Centric State}
\label{sec:plausible-latent}

The central idea of this stage is that the predicted quantity need not be an image. A model can instead predict a compact state that carries entities, relations, or task-relevant features, and can be trained without reconstructing every pixel.

Contrastively-trained Structured World Models (C-SWMs)~\cite{Kipf2019arxiv2247} are representative of this move. They structure each state embedding as a set of object representations with relations modeled by a graph network, and train the transition model contrastively rather than by reconstruction. Objects are discovered from raw observations as part of learning, so interaction structure becomes the predictive target. SlotFormer~\cite{Wu2022arxiv5861} extends the same state with transformer dynamics. It reasons autoregressively over learned object-centric features and predicts future object states across a video clip. A third design~\cite{Nematollahi2020hindsight} learns structured dynamics models in hindsight, without supervision, from recorded physical interaction.

A second line drops explicit object decomposition and keeps only the principle of predicting in embedding space. The joint-embedding predictive architecture~\cite{Assran2023ijepa} states this principle for images. V-JEPA~\cite{Bardes2024arxiv8471} applies feature prediction as a stand-alone objective for video, without pixel reconstruction or a pre-trained image encoder, and V-JEPA 2.1~\cite{MurLabadia2026arxiv4482} continues this line toward dense video features. LeWorldModel~\cite{Maes2026arxiv9312} trains such an architecture end to end from raw pixels. It uses a next-embedding prediction loss with a regularizer on the latent distribution, replacing the auxiliary supervision and moving averages that earlier variants used to avoid collapse.

This stage established that a predictive state can be far smaller than an image and still support forward prediction. It leaves the location of the predicted content unresolved. A slot index is not a position in the scene, so a drifting wall and a lost object are not separately readable.

\subsection{Metric Scene State: Occupancy, Points, and Rendering}
\label{sec:plausible-metric}

Here the question is no longer whether the state is compact, but whether its error can be measured in physical units. These methods express the future in a shared metric frame, so occupancy, depth, or displacement becomes the reported quantity.

OccWorld~\cite{Zheng2023arxiv6038} is representative of the transition. It promotes 3D semantic occupancy from an auxiliary perception output to the predicted world state. Ego motion and the evolution of the surrounding scene are then forecast in that same space. MUVO~\cite{Bogdoll2023arxiv1762} fuses camera and LiDAR history into a geometric voxel representation and studies sensor fusion for occupancy prediction. DriveWorld~\cite{Min2024arxiv4390} treats driving as a four-dimensional problem and pre-trains a memory state-space model over multi-camera video. BEVWorld~\cite{Zhang2024arxiv5679} maps multimodal sensor input into one bird's-eye-view latent space, and its decoder reconstructs LiDAR and surround images through ray-casting rendering. Within such a fixed frame, longer horizons become a generation problem. DynamicCity~\cite{Bian2024arxiv8084} compresses dynamic 4D scenes into a compact plane representation for large-scale occupancy generation. OccSora~\cite{Wang2024arxiv0337} replaces autoregressive next-token prediction with diffusion over tokens from a 4D scene tokenizer.

A related group changes the geometric primitive~\cite{song2024gvkfgaussianvoxelkernel}. ViDAR~\cite{Yang2023arxiv7655} forecasts future point clouds from visual input and uses that task for pre-training, coupling semantics, 3D structure, and time in one objective. RenderWorld~\cite{Yan2024arxiv1356} produces 3D occupancy labels with a self-supervised Gaussian-based module and represents scenes with Gaussian splatting for rendering. GaussianWorld~\cite{Zuo2024arxiv0373} reformulates 3D occupancy prediction as 4D occupancy forecasting conditioned on the current sensor input. It decomposes scene evolution into three factors: ego-motion alignment of the static scene, local movement of dynamic objects, and completion of newly observed regions. A further variant constrains the representation instead of the primitive. Geometry Forcing~\cite{Wu2025arxiv7982} aligns the intermediate representations of a video diffusion model with features from a geometric foundation model, which makes 3D structure an explicit training target. Geometry-aware 4D video generation~\cite{Liu2025arxiv1099} supervises training with cross-view point-map alignment, so that several views share one 3D scene representation. Aether~\cite{Team2025arxiv8945} jointly optimizes 4D dynamic reconstruction, action-conditioned video prediction, and goal-conditioned visual planning in one framework, so the geometry used for prediction is the geometry the model recovers.

The same principle has moved from driving to embodied manipulation and indoor scenes. TesserAct~\cite{Zhen2025arxiv0995} extends the predicted state from RGB video to RGB, depth, and normal signals over time. PointWorld~\cite{Huang2026arxiv3782} represents state and action together as 3D point flows and forecasts per-pixel displacements in 3D from RGB-D input and low-level commands. Occupancy World Model for Robots~\cite{Zhang2025arxiv5512} carries occupancy forecasting into indoor robot scenes.

This stage established a predicted state whose error is measurable in meters, occupancy, or point displacement. It also made cross-view agreement a training constraint rather than a preference~\cite{cheng2025graphguidedscenereconstructionimages,
cheng2025reggsunposedsparseviews,
cheng2025unposed3dgsreconstructionprobabilistic}. Two limits remain. The cost of the state grows with spatial extent and rollout length. Agreement between nearby views also says nothing about what happens when the sensor leaves a region and returns.

\subsection{Physics-Grounded Dynamics and Interaction Fidelity}
\label{sec:plausible-physics}

The third stage constrains how the state may change. A predicted scene can be metrically consistent and still show an object that falls through its support or passes through a wall. The variables at issue are support, collision, contact, friction, and material response, because they decide whether an interaction can occur. Agreement on these variables is what this stage calls interaction fidelity.

Benchmarks made this question separable from appearance. IntPhys~\cite{Riochet2018arxiv7616} provides a framework and benchmark for visual intuitive physics reasoning. PHYRE~\cite{Bakhtin2019arxiv5656} adds a benchmark for physical reasoning. CLEVRER~\cite{Yi2019arxiv1442} targets video representation and reasoning about collision events, and Physion~\cite{Bear2021arxiv8261} evaluates physical prediction from vision in humans and machines. These suites ask a model to judge or predict an outcome, which is why they can score event plausibility separately from frame quality.

PhyDNet~\cite{Guen2020arxiv1460} is representative of building the constraint into the predictor. It separates a branch governed by partial differential equations, realized by a recurrent physical cell, from a second branch that absorbs unknown complementary factors. OP3~\cite{Veerapaneni2019arxiv2827} encodes a different bias: it represents a scene as entities processed symmetrically under shared local dynamics, and acquires those entity representations from raw visual observation without supervision. The two designs differ in their
inductive biases, one based on equations and the other on objects.

Later systems write these constraints into generation rather than into the transition equations. DrivePhysica~\cite{Yang2024arxiv8410} constrains multi-view driving generation to respect relative and absolute motion, occlusion and spatial relations, and temporal consistency. FOLIAGE~\cite{Liu2025arxiv3173} models unbounded surface growth from images, mesh connectivity, and point clouds, advancing its latent state under physical control actions. ParticleFormer~\cite{Huang2025arxiv3126} predicts point-cloud dynamics for multi-object and multi-material manipulation, trained with a reconstruction loss on both global and local dynamics features. RoboScape~\cite{Shang2025arxiv3135} learns RGB generation jointly with physics-informed tasks, including temporal depth prediction, to improve geometry in contact-rich scenes. PhysWorld~\cite{Mao2025arxiv7416} reconstructs a physical world model from generated video, addressing the gap between photorealistic motion and physically executable behavior. WoW~\cite{Chi2025arxiv2642} argues that passive observation alone can limit physical understanding and trains a generative world model on large-scale robot interaction instead. ABot-PhysWorld~\cite{Chen2026arxiv3376} pursues the same target through physics alignment inside a manipulation world foundation model. The Cosmos platform~\cite{NVIDIA2025arxiv3575} supplies pre-trained world foundation models, video tokenizers, and post-training recipes, so a general-purpose model can be fine-tuned into a customized one for a physical AI setup.

This stage established that feasibility can be written into the predictor and scored separately from appearance. It left open whether that feasibility holds beyond a short clip.

\subsection{Persistent Memory and Long-Horizon Stability}
\label{sec:plausible-memory}

The last stage asks whether the state persists. The question is whether the same objects, layout, and map are still present at the end of a rollout, rather than how many frames the model can emit.

Persistent embodied world models~\cite{Zhou2025arxiv5495} are representative of this requirement. They add an explicit memory of previously generated content to retain scene content beyond the current observation. Video world models with long-term spatial memory~\cite{Wu2025arxiv5284} take a geometry-grounded route. They store and retrieve content through a persistent spatial memory, and use datasets curated so that revisits can be trained and evaluated. InfinityDrive~\cite{Guo2024arxiv1522} combines memory injection, memory retention, and an adaptive memory-curve loss for minute-scale driving video.

A second group extends the horizon architecturally. Long-context state-space video world models~\cite{Po2025arxiv0171} use block-wise state-space scanning to hold temporal context without the cost of attention over long sequences. LongDWM~\cite{Wang2025arxiv1546} addresses the gap between training on short clips and rolling out long videos through cross-granularity distillation. WorldDreamer~\cite{Wang2024arxiv9985} instead pursues generality rather than duration, framing world modeling as masked visual-token prediction across scenarios without a memory mechanism.

This stage established two distinct mechanisms: retrieval of stored content, and architectures that carry state further at bounded cost. A longer video demonstrates an extended rollout horizon; preserving object identity and location across those frames demonstrates state consistency. The test that distinguishes them is whether recovered content is the same object in the same place, rather than a new image that resembles the old one.

\begin{takeaway}
Plausibility requires a world state that persists and can be tested, not sharp frames. The field moved the predictive target off pixels onto latent and object-centric states, then into a metric frame where error is measured in meters, occupancy, or point displacement. It then added constraints on how that state may evolve, and finally the requirement that it survive a long rollout. Each step widened what an evaluator can test, from feature agreement to physical feasibility to persistence.
\end{takeaway}

\section{Controllable World Models}
\label{sec:controllable}

Controllability is the second capability level on the ladder. A \Clevel{} model must predict how the state changes when the agent intervenes, and not only how the scene continues on its own. The intervention may be a motor command, an ego trajectory, a latent action, a language instruction, a camera path, an object path, or a contact force. Conditioning a predictor on such a signal is not the same as obeying it. A model can accept a command, ignore it, and still return a smooth and realistic rollout. The claim therefore rests on a comparison. The requested change and the realized change must be readable in the same units, while unrelated state evolution remains consistent across matched rollouts.

\begin{figure}
    \centering
    \includegraphics[width=1.0\linewidth]{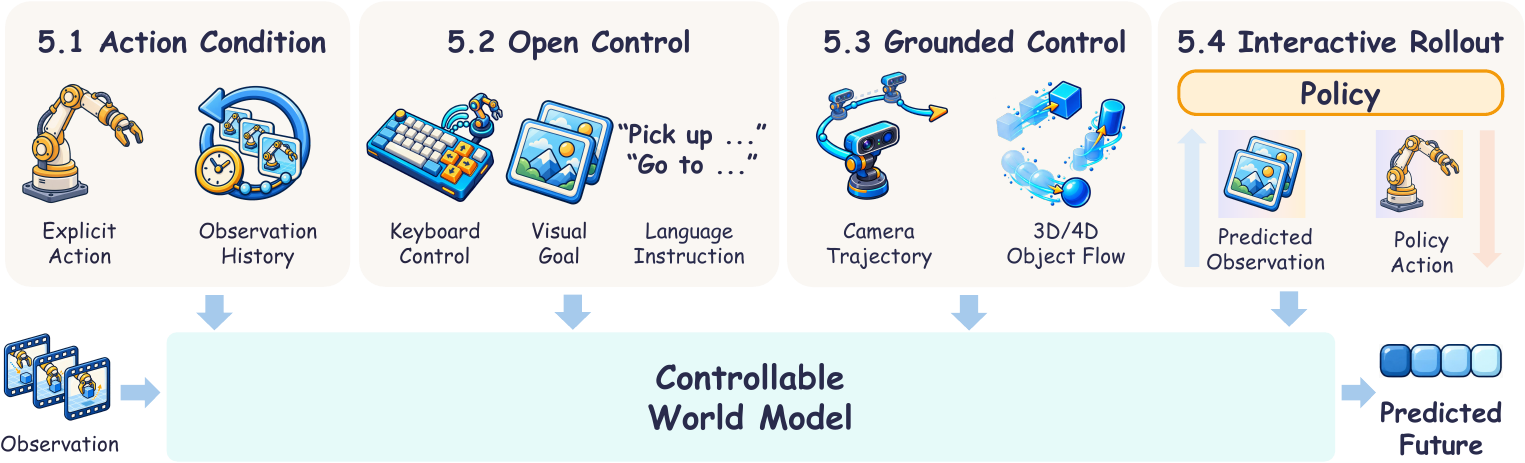}
    \caption{\textbf{Controllable World Model.} Starting from observations, control moves from fixed action conditioning with explicit commands and history (Section~\ref{sec:controllable-action}) to open, latent, and language-specified interventions such as keyboard inputs, target images, and instructions (Section~\ref{sec:controllable-open}), then to grounded controls expressed in measurable geometric or physical variables, such as camera trajectories (Section~\ref{sec:controllable-structure}), and finally to policy-driven interactive rollouts and models that couple future prediction with action generation (Section~\ref{sec:controllable-joint}). These stages progressively widen the control set, make intervention fidelity testable, and connect future prediction with action generation. The progression is conceptual rather than chronological, and the illustrated controls are examples rather than components required in every model. A Controllable claim requires the realized change to match the requested intervention while unrelated state evolution remains consistent across matched rollouts.} \label{fig:controllable-overview}
\end{figure}

This section is organized by a single question: what specifies the intervention? We group the literature into four stages of progression, as illustrated in Fig.~\ref{fig:controllable-overview}. The first is action-conditioned prediction under a fixed motor or ego command. The second widens the control set to open, latent, and language-specified interventions. The third states the intervention in a variable that an external sensor or an annotation can measure. The fourth covers policy-driven rollouts and models that couple future prediction with action generation.

\subsection{Action-Conditioned Prediction}
\label{sec:controllable-action}

The first stage keeps the control set fixed and small. The model receives the agent's own command, such as a motor action, an end-effector trajectory, or an ego driving action, and returns the observation that follows. The central idea is that the action enters the predictor as an input variable rather than as a quantity inferred from the video.

Stochastic Adversarial Video Prediction (SAVP)~\cite{Lee2018arxiv1523} is representative of the early form of this stage. It joins latent-variable and adversarial training, so that the range of possible futures is covered instead of averaged into one blurry frame. That property matters for control, because the response to a command is rarely unique. FitVid~\cite{Babaeizadeh2021arxiv3195} scales the same pixel-level interface and reports where it breaks. Existing predictors did well on narrow benchmarks and poorly on real-life data, and a model with enough capacity to close that gap overfits the benchmarks instead. PACT~\cite{Bonatti2022arxiv1133} moves the target off pixels. It predicts states and actions autoregressively as tokens, and reuses that representation as a starting point for several downstream tasks on the same robot. IRASim~\cite{Zhu2025irasim} is the clearest manipulation instance, treating rollout as trajectory-to-video generation with frame-level action conditioning inside each transformer block. Navigation World Models~\cite{Bar2025nwm} apply action-conditioned future prediction to the navigation setting.

Driving is a natural setting for this stage, because the ego action is low-dimensional and logged at scale. GAIA-1~\cite{Hu2023arxiv7080} casts driving prediction as a generative world model. DriveDreamer~\cite{Wang2023arxiv9777} learns structured traffic constraints in a first training stage and future anticipation in a second, which ties controllable generation to real driving scenes. Vista~\cite{Gao2024arxiv7398} exposes a graded control set that ranges from a high-level command or goal point to a trajectory, a steering angle, and a speed. Copilot4D~\cite{Zhang2023arxiv1017} already anticipates the next stage, since it learns a driving world model in an unsupervised manner. It tokenizes sensor observations with a VQ-VAE and predicts future tokens by discrete diffusion. ACT-Bench~\cite{Arai2024arxiv5337} closes the stage by making action fidelity a separate measurement. It pairs short context videos with future trajectory data and asks whether the generated future follows the instruction.

This stage established the conditioning interface and separated scene realism from command compliance. It left the control set tied to logged action labels of a single embodiment.

\subsection{Open, Latent, and Language-Specified Interventions}
\label{sec:controllable-open}

The second stage widens the control set. Here the central question is no longer whether a model responds to a labeled motor command, but whether an intervention can be specified when no such label exists. Two routes appeared. One recovers a control variable from unlabeled video, and the other states the intervention in language or in a goal.

Genie~\cite{Bruce2024arxiv5391} is representative of the first route. It combines a video tokenizer, an autoregressive dynamics model, and a latent action model, so a user can act frame by frame in an environment learned from unlabeled internet video. LAPA~\cite{Ye2024arxiv1758} quantizes discrete latent actions between frames, and then maps them to robot actions using a small amount of labeled manipulation data. AdaWorld~\cite{Gao2025arxiv8938} extracts latent actions during pre-training and conditions an autoregressive world model on them, which supports adaptation to environments with heterogeneous action spaces. LDA-1B~\cite{Lyu2026arxiv2215} scales the same idea by ingesting heterogeneous embodied data into one latent dynamics action model.

A parallel group keeps the action explicit and opens the environment or the vocabulary instead. UniSim~\cite{Yang2023arxiv6114} orchestrates datasets that are rich along different axes, combining abundant objects, densely sampled robot actions, and diverse navigation motion in one interactive simulator. MineWorld~\cite{Guo2025arxiv8388} interleaves image tokens and action tokens in a game setting, and decodes them in parallel so that a human can interact in real time. Yume~\cite{Mao2025arxiv7744} quantizes camera motion, so a world generated from one image can be explored with keyboard input. Pandora~\cite{Xiang2024arxiv9455} accepts free-text actions by joining a pretrained language model with a pretrained video model.

Language also serves as the specification in manipulation. RoboDreamer~\cite{Zhou2024arxiv2377} parses an instruction into lower-level primitives and factorizes generation over them, so an unseen instruction can be composed from seen components. ManipDreamer~\cite{Li2025arxiv6464} represents the instruction as an action tree, which models relations between primitives instead of treating them as independent, and adds depth and semantic guidance. The Object-Centric World Model~\cite{Jeong2025arxiv6170} predicts future states in a slot representation conditioned on the instruction, which keeps the predicted variable compact. Goal-VLA~\cite{Chen2025arxiv3919} moves the specification one step further. An image-generative vision-language model produces the desired goal state, and the target object pose is derived from it.

This stage reduced the dependence on labeled actions. It also moved the command into units that are internal to the model, so a latent or linguistic intervention cannot be compared directly with the realized change.

\subsection{Structure-Grounded Control: Geometry, Trajectory, and Contact}
\label{sec:controllable-structure}

The third stage answers the measurement problem that the previous one created. The intervention is stated in a variable that an external sensor or an annotation can measure. Examples are a camera pose, an occupancy grid, an object path, a joint configuration, and a contact force. The comparison then needs no human rating.

MagicDrive~\cite{Gao2023arxiv2601} is representative for geometry. It conditions street-view generation on camera poses, road maps, 3D bounding boxes, and text, and adds cross-view attention for consistency across cameras. DOME~\cite{Gu2024arxiv0429} keeps the predicted state in a native 3D form by forecasting future occupancy frames, and introduces trajectory resampling for fine-grained control.

In manipulation the measurable variable is usually a path. Mask2IV~\cite{Li2025arxiv3135b} predicts interaction trajectories for the actor and the object in a first stage, then generates video conditioned on them, which removes the need for dense user-supplied masks. RoboMaster~\cite{Fu2026robomaster} decomposes an interaction into phases and controls the coupled motion of the arm and the manipulated object along a trajectory. In 3DFlowAction~\cite{Zhi2025arxiv6199}, the commanded variable is the future 3D flow of the interacting object, which is embodiment-agnostic and therefore shared between human and robot data. GAF~\cite{Chai2025arxiv4135} extends 3D Gaussian splatting with learnable motion attributes, so current scene reconstruction, future frames, and action-aware motion are read from one 4D field. In 3D-VLA~\cite{Zhen2024arxiv9631}, the target is stated as a goal image and a goal point cloud, produced by embodied diffusion models aligned to a 3D language model.

The same logic extends to physical variables that are not directly observed in an image, such as mass, friction, and contact force. DreMa~\cite{Barcellona2024arxiv4957} builds a digital twin by joining Gaussian splatting with a physics simulator, so novel object configurations can be imagined and the consequences of a robot action predicted. DexSim2Real$^2$~\cite{Jiang2024arxiv8750} constructs an explicit model of an unseen articulated object through active interaction, and plans toward different goals with sampling-based model predictive control. PIN-WM~\cite{Li2025arxiv6693} identifies 3D rigid-body dynamics from visual observation using differentiable physics and an observational loss induced by Gaussian splatting. Physically Embodied Gaussian Splatting~\cite{AbouChakra2024arxiv0788} pairs particles governed by a physics system with attached Gaussians, and converts the gap between predicted and observed images into corrections of particle positions. OmniVTA~\cite{Zheng2026arxiv9201} treats tactile signals as an active channel for modeling contact dynamics rather than as passive observations. It pairs a large visuo-tactile-action dataset for contact-rich manipulation with explicit closed-loop control.

This stage made the requested change readable in physical units. Its cost is the annotation, reconstruction, or sensing needed to obtain those units.

\subsection{Interactive Rollout and Joint World--Action Models}
\label{sec:controllable-joint}

The fourth stage changes who supplies the intervention. The command no longer comes from a logged sequence but from a policy acting inside the model. The rollout must stay faithful while that policy reacts to its own predicted observations. Joint world--action models also produce actions.

Ctrl-World~\cite{Guo2025arxiv0125} is representative of this transition. It combines joint multi-view prediction, frame-level action control, and memory-based long-horizon generation, so a generalist policy can be rolled out inside it over many steps. The token-based route is taken by iVideoGPT~\cite{Wu2024arxiv5223}, which places visual observations, actions, and rewards in one autoregressive sequence with compressive tokenization. Vid2World~\cite{Huang2025arxiv4357} converts a pretrained video diffusion model into this regime by causalizing its architecture and training objective and adding causal action guidance. DreamDojo~\cite{Gao2026arxiv6949} builds a comparable interface from large-scale human video with continuous latent actions, then post-trains it for a target embodiment. Mem-World~\cite{Zheng2026arxiv8960} targets the forgetting and hallucination that follow from end-effector occlusion and rapid wrist-camera motion. It augments a multi-view action-conditioned model with a memory of earlier frames and a rule for retrieving informative history. EnerVerse~\cite{Huang2025arxiv1895} predicts a multi-view future space from instructions with chunk-wise autoregressive video diffusion, and translates the resulting 4D representation into robot actions. EnerVerse-AC~\cite{Jiang2025arxiv9723} formulates an action-conditional multi-view generator that also serves as an evaluator of robot policies. Policy evaluation is the subject of Section~\ref{sec:actionable}, and only rollout behavior is treated here.

The second half of the stage couples prediction with control. GR-1~\cite{Wu2023arxiv3139} pre-trains on large-scale video, then predicts robot actions and future images end to end from an instruction, an image history, and a state history. GR-2~\cite{Cheang2024arxiv6158} keeps that recipe and fine-tunes one model for video generation and action prediction together. In Video Prediction Policy~\cite{Hu2024arxiv4803} the two components stay separate, and an implicit inverse dynamics model is learned on the future representations inside a video diffusion model. Prediction with Action (PAD)~\cite{Guo2024arxiv8179} places future images and robot actions in a single denoising process, which allows co-training on demonstrations and on large video datasets. In Unified Video Action~\cite{Li2025arxiv0200}, a joint video--action latent decouples the two decoders, so action inference does not require generating video. Unified World Models~\cite{Zhu2025arxiv2792} give each modality its own diffusion timestep, so one transformer can act as a policy, a forward model, an inverse model, or a video generator. WorldVLA~\cite{Cen2025arxiv1539} interleaves action tokens and image tokens autoregressively, and masks prior actions to limit error propagation inside a chunk. DyWA~\cite{Lyu2025arxiv6806} predicts future states while adapting to dynamics variation inferred from the trajectory history, which couples geometry, state, physics, and action for non-prehensile manipulation. EVA~\cite{Wang2026arxiv7808} post-trains a video world model with inverse-dynamics rewards to encourage alignment with executable action sequences.

This stage connects action-conditioned prediction with action generation, although individual methods support different directions of this interface. It left open how far a rollout stays faithful once a policy, rather than a dataset, chooses the actions.

\begin{takeaway}
Controllability is intervention fidelity, not conditioning. The literature first widened the intervention from a labeled motor or ego command to latent, open, and language-specified controls. It then restated the intervention in geometric and physical variables that can be measured directly. Current approaches support policy-driven rollouts or couple future prediction with action generation.
\end{takeaway}

\section{Actionable World Models}
\label{sec:actionable}

\begin{figure}[t]
    \centering
    \includegraphics[width=1.0\linewidth]{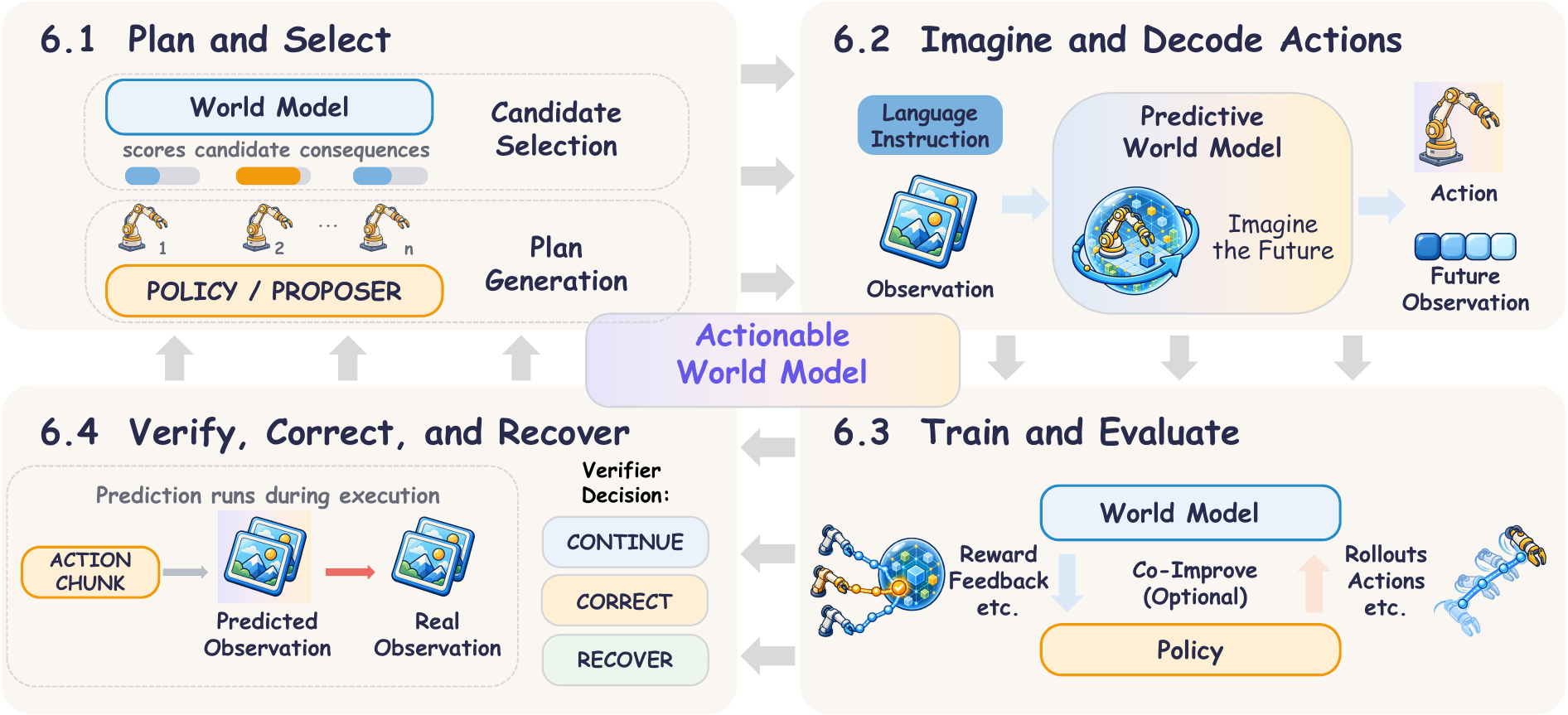}
    \caption{\textbf{Actionable world models.} Each quadrant is one stage of Section 6, the central card the capability being established, and the grey connectors the clockwise reading order rather than runtime data flow. In Section~\ref{sec:actionable-planning} a proposer supplies candidate behavior and predicted consequences support selection; in Section~\ref{sec:actionable-policy} imagined futures or predictive features support action decoding; in Section~\ref{sec:actionable-simulation} a learned simulator supplies rollouts and feedback for policy training or evaluation; and in Section~\ref{sec:actionable-verification} predicted and observed execution are compared to continue, correct, or recover. The stages represent alternative placements of prediction in the decision process rather than a mandatory closed pipeline, and an Actionable claim requires a measured downstream gain over a matched baseline.}
    \label{fig:overview-actionable}
\end{figure}

An \Alevel{} world model changes a decision or a learning update and improves a measured outcome. The criterion stated in Equation~\ref{eq:actionability} has two parts. The prediction must alter what the system does, and the altered behavior must be better than a matched alternative that does not use it. Plausibility requires consistent state predictions and controllability requires correct responses to interventions; downstream utility establishes neither on its own. Prediction errors matter when they distort distinctions that the decision maker needs. What must be correct depends on who consumes the prediction. A planner, an action decoder, a learner, an evaluator, and a runtime verifier each depend on a different property of the predicted future.

We organize the literature into four functional stages, as depicted in Fig.~\ref{fig:overview-actionable}: value-based selection outside the policy, action decoding from imagination inside it, learned simulators for policy training and evaluation, and runtime verification and recovery. These stages differ in where prediction enters the decision process and in the outcome measures and matched comparisons used to assess its utility.



\subsection{Planning, Search, and Value-Based Selection}
\label{sec:actionable-planning}

In the first stage the world model sits outside the policy. Some other mechanism proposes candidate behavior, the model predicts the consequence of each candidate, and a score orders them. The central design choice is what the search runs over: raw action sequences, subgoals, poses, or symbolic states.

Video Language Planning~\cite{Du2023arxiv0625} is representative of this arrangement. It runs a tree search in which vision--language models serve as policies and value functions and a text-to-video model serves as the dynamics model. The result is a long video plan for a long-horizon instruction. Prediction expands and scores branches; it does not emit the action. Large Video Planner~\cite{Chen2025arxiv5840} extends the same video-plan approach toward generalizable robot control.

An earlier line had already used learned dynamics for selection under constraints. Safe MPC~\cite{Koller2019arxiv2189} builds confidence intervals on predicted trajectories from a statistical model, and admits only inputs that keep the system safe with high probability while it learns. VIPER~\cite{Escontrela2023arxiv4343} moves the learned component into the score itself: an autoregressive video prediction model trained on expert videos supplies likelihoods that replace a hand-specified task reward for a reinforcement learning agent. DINO-WM~\cite{Zhou2024arxiv4983} drops pixel reconstruction from the rollout, modeling visual dynamics on pre-trained DINOv2 patch features. It optimizes behavior at test time toward a given goal.

Another group of systems changes what proposes the candidates and how the search is spent. Drive-WM~\cite{Wang2023arxiv7918} generates multiview futures for distinct driving maneuvers so that an end-to-end planner can compare them. WorldPlanner~\cite{Khorrambakht2025arxiv3077} learns an action-conditioned visual world model and a diffusion action sampler from unstructured play data, then combines Monte Carlo tree search with model-predictive control. STORM~\cite{Lin2025arxiv8477} lets a diffusion policy propose candidates, simulates their visual and reward outcomes with a generative video model, and refines the plan by lookahead search. GHIL-Glue~\cite{Hatch2025ghilglue} moves selection up one level, filtering generated subgoal images before a low-level policy consumes them. Inference-Time Enhancement~\cite{Qi2026ralitps} keeps the policy fixed and uses predictive world modeling only to improve what it emits at test time. Reimagination with Observation Intervention~\cite{Chen2025arxiv6565} instead edits, at test time, the observation that the model conditions on, so that unfamiliar distractors do not corrupt the predictions used by visual model-predictive control.

A parallel direction reduces the cost of search by coarsening the state that is searched over. From Pixels to Predicates~\cite{Athalye2026ral} learns symbolic predicates from images with pre-trained vision--language models, and Exopredicator~\cite{Liang2026exopredicator} learns abstract models of dynamic worlds for planning. StructVLA~\cite{Jin2026arxiv2553} replaces dense future frames with an explicit structured plan that retains kinematic grounding. World Action Planner~\cite{Zhang2026arxiv7599} has a vision--language model propose an initial plan, then refines it by search over pose-image conditioned rollouts. The World-Value-Action model~\cite{Li2026arxiv4732} pushes further by learning a value over latent future trajectories, so that comparison happens without decoding a rollout. EvolvingAgent~\cite{Feng2025arxiv5907} instead makes the model non-stationary for long-horizon open-world tasks. An LLM task planner and a world-model-guided controller are paired with a curriculum-based reflector that selects experiences for continual model updates.

This stage established that a world model can act as the scoring function for behavior that something else proposes, and that the score can be learned rather than written by hand. It also exposed the cost structure of that arrangement. Every candidate consumes a rollout, so the proposal count and the number of model calls bound what can run inside a control period.

\subsection{Policy Learning and Action Decoding from Imagination}
\label{sec:actionable-policy}

The second stage moves prediction inside the policy. The question is no longer which candidate to select, but whether a predictive objective, or a predicted future itself, can produce the action.

ContextWM~\cite{Wu2023arxiv8499} is representative of the earliest form. It pre-trains a world model on in-the-wild videos so that background and appearance variation do not obscure the shared dynamics, and transfers the result to downstream visual control. Prediction here is unsupervised pre-training of the world model itself, which improves the sample efficiency of downstream model-based reinforcement learning.

A second form decodes the action directly from a generated future. Dreamitate~\cite{Liang2024arxiv6862} fine-tunes a video diffusion model on human demonstrations with a common tool, synthesizes an execution in a novel scene, and uses that synthesized execution to control the robot. NovaFlow~\cite{Li2026novaflow} extracts actionable flow from generated video for zero-shot manipulation, and TC-IDM~\cite{Mi2026arxiv8323} grounds video generation into executable zero-shot robot motion. Video Generators are Robot Policies~\cite{Liang2025arxiv0795} makes the same identification directly, and Grounding Video Models to Actions~\cite{Luo2025grounding} recovers the missing action mapping through goal-conditioned exploration.

A third form keeps the prediction internal. VidMan~\cite{Wen2024arxiv9153} uses a two-stage procedure that exploits the implicit dynamics of a video diffusion model and then applies them to manipulation. FLARE~\cite{Zheng2025arxiv5659} aligns features of a diffusion transformer policy with latent embeddings of future observations, adding only a few tokens to a standard vision--language--action model. JEPA-VLA~\cite{Miao2026arxiv1832} and FRAPPE~\cite{Zhao2026arxiv7259} follow the same route through predictive embeddings and alignment with multiple future representations, while UP-VLA~\cite{Zhang2025arxiv8867} trains understanding and prediction in one embodied model. In these systems the future is never displayed; it constrains the representation from which actions are read out. X-MOBILITY~\cite{Liu2024arxiv7491} extends this arrangement to end-to-end navigation. It decouples world modeling from the action policy, so that off-policy data trains the latent dynamics and on-policy data trains the control.

World--action models make both streams explicit outputs of one backbone. DiT4DiT~\cite{Ma2026arxiv0448} couples a video diffusion transformer with an action diffusion transformer in a cascade. DreamZero~\cite{Ye2026arxiv5922} builds a world--action model on a pre-trained video diffusion backbone and learns from heterogeneous robot data by jointly modeling video and action. WSA$_1$~\cite{Jiang2026arxiv3941} adds a 3D-centric state so that world evolution, spatial structure, and action share one representation, and DriveWorld-VLA~\cite{jia2026arxiv6521} unifies latent world modeling with planning for driving.

The cost of rendering a future then became the design variable. LaWAM~\cite{Chen2026arxiv5768} exposes compact latent visual subgoals instead of reconstructed video. ForeWAM~\cite{Huang2026arxiv1605} supplies predictive context to a direct-policy model without decoding future video. AHEAD~\cite{Syed2026arxiv2486} forecasts future patch tokens in the feature space of a frozen policy, using per-token velocity and acceleration from optical flow. The policy then acts on where a moving object will be rather than where it was observed. AHA-WAM~\cite{Cai2026arxiv9811} decouples the temporal rhythm of world prediction from that of action execution. Fast-WAM~\cite{Yuan2026arxiv6666} asks whether explicit imagination is needed at test time, or whether the benefit comes from video modeling during training. The $\tau_0$ world model~\cite{Zhou2026arxiv1027} keeps both options open, offering a joint video--action interface and an action-conditioned simulator that rolls out candidate action chunks.

These methods indicate that predictive structure can benefit action generation without explicit future decoding at run time; both predictive reliability and latency constrain its use. The next stage considers the model as an environment for policy training and evaluation.

\subsection{Learned Simulators for Policy Training and Evaluation}
\label{sec:actionable-simulation}

In the third stage the model becomes the environment. Policies act inside it, are updated from its outcomes, and are ranked by its results. The requirement changes from accuracy on logged transitions to reliability along the states that the policy itself visits.

SafeDreamer~\cite{Huang2023arxiv7176} is representative of the training use. It plans and learns inside imagined trajectories while treating a safety cost as an explicit constraint through Lagrangian methods. Think2Drive~\cite{Li2024arxiv6720} trains a driving planner inside a latent world model rather than in the simulator directly. VLM-SAFE~\cite{Qu2025arxiv6377} organizes offline safe reinforcement learning around an observe, imagine, evaluate, and act cycle, with a vision--language model supplying the semantics of risk. DriveArena~\cite{Yang2024arxiv0415} closes the loop at the platform level, pairing a traffic manager with a generative world model so that a driving agent perceives generated images and returns trajectories.

In robotics the same use appears with an explicit interface between the simulator and the learner. Robotic World Model~\cite{Li2025arxiv0100} presents a neural simulator with dual-autoregressive prediction and a policy-optimization procedure that trains in imagination and deploys on the system. LUMOS~\cite{Nematollahi2025arxiv0370} practices language-conditioned skills over many long rollouts in the latent space of a learned world model and transfers them to a real robot without further tuning. In general control settings, SENSEI~\cite{Sancaktar2025arxiv1584} broadens what such a latent environment covers, distilling an interestingness reward from vision--language model annotations to drive semantic exploration. Two further methods target the gap to the real system. EmbodieDreamer~\cite{Wang2025arxiv5198} aligns physical parameters and appearance for real-to-sim-to-real transfer, and ReDRAW~\cite{Lanier2025arxiv2252} calibrates a simulation-pretrained world model to a target environment through residual corrections of latent-state dynamics.

A further group places vision--language--action and diffusion policies inside learned world models. World-Env~\cite{Xiao2025arxiv4948} uses the world model as a virtual environment for post-training a vision--language--action policy, and WMPO~\cite{Zhu2026wmpo} formulates policy optimization against such a model. VLA-RFT~\cite{Li2025arxiv0406} performs reinforcement fine-tuning with verified rewards computed inside a world simulator. DiWA~\cite{Chandra2025diwa} adapts a diffusion policy with a world model, and Prophesying~\cite{Zhang2025arxiv0633} reinforces an action policy against predicted outcomes. VLAW~\cite{Guo2026arxiv2063} makes the coupling explicit by improving the policy and the world model in alternation; Section~\ref{sec:loops} examines what repeated updates of this kind require.

The second use of the same machinery is evaluation. SIMPLER~\cite{Li2024arxiv5941} is the reference point: it identifies control and visual disparities between real and simulated settings and treats agreement with real-robot evaluation as the criterion a simulated evaluation must meet. WorldEval~\cite{Li2025arxiv9017} and WorldGym~\cite{Quevedo2025arxiv0613} replace the hand-built simulator with an action-conditioned video model, the latter scoring Monte Carlo rollouts with a vision--language model and starting from real initial frames. RoboWorld~\cite{Jeon2026arxiv1060} addresses the two practical obstacles of that route, unreliable long rollouts and slow inference, with anchored context and progress-aware scoring. Genie Envisioner~\cite{Liao2025arxiv5635} integrates policy learning, evaluation, and simulation in one video-generative platform, and the Gemini Robotics team~\cite{GeminiRobotics2025arxiv0675} reports evaluation of its policies inside a Veo world simulator.

Benchmarks then ask what a simulator must satisfy before it may serve either role. WorldSimBench~\cite{Qin2024arxiv8072} organizes predictive-model functions into a hierarchy and pairs perceptual assessment with a manipulative one, and EWMBench~\cite{Yue2025arxiv9694} separates scene, motion, and semantic quality. WorldArena~\cite{Shang2026arxiv8971} adds functional utility in decision making next to perceptual quality, and RoboWM-Bench~\cite{Jiang2026arxiv9092} tests whether predicted interactions are executable by a robot. MiraBench~\cite{Yang2026arxiv9360} decomposes action-conditioned reliability into physics adherence, faithfulness to commanded actions, and calibration when actions should not succeed.

This stage established that one learned model can host both the policy update and the audit of the result. It also produced the difficulty that defines the stage. A model that trains a policy cannot by itself certify it, because the policy is optimized against the same errors the audit must detect.

\subsection{Runtime Verification, Correction, and Recovery}
\label{sec:actionable-verification}

The fourth stage runs prediction during execution. Two execution-specific conditions matter. The signal must arrive inside the control period, and the system must act on it. The literature separates into four claims of increasing strength. They are diagnosis of what went wrong, detection while the episode runs, correction of the actions still to be issued, and recovery of a state from which the task can continue.

Diagnosis comes first because it establishes which discrepancies carry information about failure. RoboFAC~\cite{Ye2025arxiv2224} builds a failure-centric dataset of erroneous manipulation trajectories with categorized failure types and question--answer supervision for analysis and correction. ViFailback~\cite{Zeng2025arxiv2787} annotates failures with explicit visual symbols and returns both textual and visual correction guidance, so that a diagnosis is expressed in a form a controller can use. PhysicalAgent~\cite{Lykov2025arxiv3903} closes a coarse loop around this idea by generating candidate trajectory videos, executing them, and re-planning after a failure. ST-WAM~\cite{Wang2026arxiv8993} diagnoses the predictor instead of the policy, identifying futures that hallucinate training-domain content under visual shift and selecting features that stay stable across that shift.

Runtime verifiers compare the predicted and the observed transition at the rate of execution. CheckVLA~\cite{Liu2026arxiv6789} checks a committed action chunk against a separately trained, frozen action-conditioned world model, providing a reference whose parameters remain fixed during verification. Foresight~\cite{Zhang2026arxiv3085} monitors latent world-model representations over long-horizon trajectories and is trained from final success or failure labels only, avoiding dense annotation of failure onset. FoMo-FD~\cite{Huang2026arxiv7511} learns nominal short-horizon dynamics with an action-conditioned flow-matching model for surgical imitation policies. Two systems make the alarm statistical. Foundational World Models for Bimanual Failure Detection~\cite{Ward2026arxiv6987} run a probabilistic model in the latent space of a pre-trained tokenizer, and use its uncertainty as a non-conformity score. World Model Failure Classification~\cite{Ho2026arxiv6182} applies conformal thresholds to separate success, known failure, and out-of-distribution cases in autonomous inspection. Other verifiers change what is compared rather than how. ActFovea~\cite{Yu2026arxiv9169} restricts the comparison to action-conditioned foveated regions that retain contact-relevant areas. FAWAM~\cite{He2026arxiv8555} predicts future end-effector wrenches, so that contact errors invisible in pixels become residuals.

Correction is the first claim that changes behavior rather than reporting on it. Because generative policies commit to a chunk, the decision is when to stop trusting it. When to Trust Imagination~\cite{Wang2026arxiv6222} formulates adaptive execution as future--reality verification, executing longer while the imagined future holds and replanning when it does not. TempoWAM~\cite{Ye2026arxiv9492} makes the same decision from an online estimate of task progress rather than a step count. VLA-Corrector~\cite{Pan2026arxiv1804} truncates stale actions once latent deviation persists and generates a corrected suffix, and AdaReP~\cite{Cheng2026arxiv3079} adapts the tolerance for reusing a cached plan from the current mismatch. Feedback World Model~\cite{An2026arxiv5705} corrects the predictor rather than the plan, updating its state with the observed next state before it guides the policy again. Two systems in this group address the delay between inference and execution directly. FutureRTC~\cite{Jiang2026arxiv4008} predicts the observations and states that will hold when a chunk is finally executed. World Action Models in Real Time~\cite{Team2026arxiv1880} compares asynchronous strategies for overlapping inference with execution on a bimanual robot.

Recovery is needed when no suffix of the current plan reaches the goal. PLanAR~\cite{Guo2026arxiv1662} defines the reasoning space of a vision--language model with object predicates, action schemas that carry preconditions and effects, and symbolic plans. See, Plan, Rewind~\cite{Dai2026arxiv9292} grounds an instruction into spatial milestones and rewinds to a recoverable state when progress stalls. CycleVLA~\cite{Ma2026arxiv2295} anticipates failure at subtask transitions and backtracks to an earlier subtask instead of correcting in place. Back to the Familiar Future (B2FF)~\cite{Shin2026arxiv9258} pre-computes a bank of familiar future states from the clean initial observation. Selecting one of them as the recovery target aims to return execution toward states familiar to the policy.

This stage established that the residual between an imagined and an observed execution is itself a control signal. Its value depends on when it arrives as much as on how accurate it is. It left a harder question open, since an alarm that fires on a correct execution has its own cost.

\begin{takeaway}
\Alevel{} world models are distinguished by measured downstream utility across different placements of prediction in the decision process. The literature moves from scoring externally proposed candidates, to generating the action, to hosting the training and ranking of policies, to correcting execution while it happens. These uses impose different requirements on latency and on the independence of the signals used to evaluate outcomes.
\end{takeaway}

\section{Four Improvement Loops}
\label{sec:loops}

The capability ladder says what a world model demonstrates. The improvement loops say how its predictions change the system over repeated interaction. Sections~\ref{sec:plausible}--\ref{sec:actionable} organize studies by the capabilities they address; this section examines update paths that may support repeated improvement.

Four loops carry that cycle, as defined in Section~\ref{sec:taxonomy}. The data loop acquires, generates, or curates experience. The reward loop turns prediction into evaluative feedback. The policy loop changes action selection or policy parameters using predicted futures. The model-self loop repairs the predictor. Table~\ref{tab:grounding-improvement-matrix} crosses the four with geometry, physics, and action grounding; Table~\ref{tab:loops} states what circulates in each loop and what settles its claim. The loops are concurrent rather than sequential, and one system often runs several at once.

Let $M_{\theta}$ denote a world model, $\pi_{\phi}$ a policy, $C_{\psi}$ a reward, value, or critic model, and $D$ an interaction buffer. A coupled iteration can be written as
\begin{equation}
\begin{aligned}
D_{k+1} &= D_k \cup \mathcal{C}(M_{\theta_k},C_{\psi_k},\pi_{\phi_k}),\\
(\theta,\psi,\phi)_{k+1} &= \mathcal{U}(D_{k+1};\theta_k,\psi_k,\phi_k),
\end{aligned}
\label{eq:loop}
\end{equation}
where $\mathcal{C}$ collects, generates, or filters experience and $\mathcal{U}$ updates any subset of the three models. The subscript $k$ in Equation~\ref{eq:loop} is what distinguishes this section from the previous three. A single application of $\mathcal{C}$ and $\mathcal{U}$ supports an \Alevel{} claim only if it improves a measured outcome over a matched alternative; a claim of sustained improvement additionally requires evaluation across repeated updates.

\begin{table}[!t]
  \centering
  \caption{%
    The four improvement loops of Section~\ref{sec:loops}.
    \emph{What circulates} names the quantity the loop passes from the
    predictor back into the system. \emph{Self-reinforcing failure} names what
    goes wrong when the loop is closed on itself.
    \emph{Settlement metric} names the measurement that decides whether
    repeated updates helped.
  }
  \label{tab:loops}
  \footnotesize
  \setlength{\tabcolsep}{3pt}
  \renewcommand{\arraystretch}{1.05}
  \begin{tabularx}{\textwidth}{
    @{}
    >{\raggedright\arraybackslash}p{0.13\textwidth}
    >{\raggedright\arraybackslash}p{0.22\textwidth}
    >{\raggedright\arraybackslash}X
    >{\raggedright\arraybackslash}X
    >{\raggedright\arraybackslash}X
    @{}
  }
    \toprule
    \textbf{Loop}
    & \textbf{Representative methods}
    & \textbf{What circulates}
    & \textbf{Self-reinforcing failure}
    & \textbf{Settlement metric} \\
    \midrule

    \textbf{Data Loop}
    (Sec.~\ref{sec:loop-data})
    & V-JEPA~2~\cite{Assran2025arxiv9985}, DreamGen~\cite{Jang2025dreamgen}, GigaWorld-0~\cite{GigaWorld2025arxiv9861}, Dream2Fix~\cite{Li2026arxiv3528}, WorldSimProbe~\cite{Co2026arxiv9298}
    & Synthesized or newly acquired transitions: coverage in the broad case,
      transitions the predictor cannot explain in the targeted case.
    & Coverage narrows toward what the current model already renders well.
    & Held-out capability per unit of real interaction, against uniform
      collection. \\

    \textbf{Reward and critic Loop}
    (Sec.~\ref{sec:loop-reward})
    & GE-Sim~2.0~\cite{Qiu2026arxiv7491}, RISE~\cite{Yang2026arxiv1075}, Self-Correcting VLA~\cite{Liu2026arxiv1633}, WCM~\cite{Fei2026arxiv9613}
    & A scalar score for an imagined or observed outcome.
    & The policy learns to satisfy the score rather than the task.
    & Agreement between predicted and realized outcomes on policy-generated actions. \\

    \textbf{Policy Loop}
    (Sec.~\ref{sec:loop-policy})
    & Cosmos Policy~\cite{Kim2026arxiv6163}, WoVR~\cite{Jiang2026arxiv3977}, World4RL~\cite{Jiang2025arxiv9080}, WorldSample~\cite{Xue2026arxiv2431}, World-in-World~\cite{Zhang2025arxiv8135}
    & Imagined rollouts used as training experience or as the planning
      environment.
    & Behavior improves inside the model and degrades on the real system.
    & Target-environment return at matched sample and compute budget. \\

    \textbf{Model-self Loop}
    (Sec.~\ref{sec:loop-model-self})
    & RLVR-World~\cite{Wu2025arxiv3934}, World Action Verifier~\cite{Liu2026arxiv1985}, World-VLA-Loop~\cite{Liu2026arxiv6508}
    & Measured prediction error, fed back into the predictor.
    & The model forgets rare dynamics and absorbs the current policy's bias.
    & Gain on the diagnosed failure and regression on a fixed audit set. \\

    \bottomrule
  \end{tabularx}
\end{table}

\subsection{Data Loop}
\label{sec:loop-data}

The data loop uses a world model to acquire, generate, or curate transitions. Failure-directed collection is one way to make those transitions more informative about geometry, physics, or action effects.

DreamGen~\cite{Jang2025dreamgen} is representative of the generative route. It adapts a video world model to a target embodiment, synthesizes neural trajectories, and recovers executable actions from them through latent action modeling or inverse dynamics. GigaWorld-0~\cite{GigaWorld2025arxiv9861} makes the data-engine view explicit, pairing a controllable video branch with a physically grounded 3D branch for large-scale embodied data synthesis. Interactive World Simulator~\cite{Wang2026arxiv8546} uses consistency models for both image decoding and latent dynamics, so rollouts stay fast and stable enough to collect demonstrations and evaluate policies inside the model. V-JEPA~2~\cite{Assran2025arxiv9985} shows how far breadth alone reaches. It pre-trains an action-free predictor on internet video, then post-trains a latent action-conditioned world model on a much smaller set of unlabeled robot video. That model plans toward image goals. These four widen coverage. What none of them does is let the predictor's own error decide which transition comes next.

Failure-oriented methods provide more targeted generation or diagnostics. Dream2Fix~\cite{Li2026arxiv3528} perturbs actions inside a generative world model to synthesize counterfactual failure rollouts paired with corrections, which success-only demonstrations never contain. A structured verification step then filters those rollouts for task validity, visual coherence, and kinematic safety. WorldSimProbe~\cite{Co2026arxiv9298} instead states a contract that any action-conditioned simulator should satisfy: supplied actions must induce the corresponding motion, and environment responses must follow from that realized motion. Its five controlled suites report where action realization degrades.

These studies illustrate targeted synthesis and diagnostics that could guide acquisition. Establishing a repeated error-driven collection loop requires further evaluation, including independent checks of synthetic transition coverage.

\subsection{Reward and Critic Loop}
\label{sec:loop-reward}

The reward loop compresses a transition into preference, return, progress, feasibility, or risk. It supplies dense supervision where task reward is sparse. It also creates an optimization target that a changing policy can exploit.

GE-Sim~2.0~\cite{Qiu2026arxiv7491} is representative of the simulator-level form. Its world judge scores generated rollouts against the task instruction, which replaces manual inspection with an automated success estimate. RISE~\cite{Yang2026arxiv1075} splits a compositional world model into a controllable dynamics model that predicts multi-view futures and a progress value model that scores them. The score is produced by the predictor itself.

Policy-coupled critics shorten the path from prediction to online optimization. Self-Correcting VLA~\cite{Liu2026arxiv1633} adds predictive heads that forecast task progress and near-term trajectory trends, then uses those sparse predictions to reshape a dense progress reward and refine actions online. WCM~\cite{Fei2026arxiv9613} trains the critic with an explicit world-modeling objective, jointly predicting future latent state and estimating value, so the estimate rests on temporal structure rather than a single frame.

These studies illustrate how predictive models can supply or augment evaluative feedback. A remaining concern is ordering: a critic that is accurate on average can still reverse the ranking of the two candidates the controller must choose between.

\subsection{Policy Loop}
\label{sec:loop-policy}

The policy loop uses the model to propose, label, filter, or rehearse behavior. Its advantage is scale, because an agent can visit more futures than it can safely execute. Its failure mode is model exploitation.

Cosmos Policy~\cite{Kim2026arxiv6163} is representative of the offline form. A single stage of post-training turns a pretrained video model into the policy, with actions encoded as latent frames and no architectural change. The same model also generates future states and values, which support planning at test time. World4RL~\cite{Jiang2025arxiv9080} refines a manipulation policy against a diffusion world model with reinforcement learning. WoVR~\cite{Jiang2026arxiv3977} extends that setting to vision--language--action policies and regulates how reinforcement learning meets imperfect imagined dynamics. It initializes rollouts from keyframes to limit error depth and co-evolves the world model with the policy it trains. GigaBrain-0.5~M*~\cite{GigaBrain2026arxiv2099} applies the same world-model-based reinforcement learning at the scale of a full vision--language--action system.

WorldSample~\cite{Xue2026arxiv2431} carries this to closed-loop reinforcement learning on real robots, generating synthetic transitions from a post-trained world model and regulating which of them enter training through sample selection and scheduling. World-in-World~\cite{Zhang2025arxiv8135} instead compares heterogeneous world models under one online planning procedure and a standardized action API, ranking them by closed-loop task success. That makes the induced behavior a criterion for choosing a predictor. Updates in this loop should be gated by model support. The gate may be soft, but the fraction of imagined experience it accepts is part of the result.

Imagined experience can support policy improvement, but a support check alone does not establish a gain in realized outcomes. A remaining question is where to set that gate, since a strict threshold discards useful imagined experience and a loose one admits states the model cannot predict.

\subsection{Model-Self Loop}
\label{sec:loop-model-self}

The model-self loop updates the predictor itself, changing predictive parameters, state, memory, or uncertainty from measured error. A better architecture is not a loop. The update must be triggered by verification results or prediction errors observed during deployment and must target a diagnosed weakness.

RLVR-World~\cite{Wu2025arxiv3934} is representative of the outcome-verified form. It optimizes a world model by reinforcement learning against the transition-prediction metric it will be judged on, computed on decoded predictions, rather than against likelihood. World Action Verifier~\cite{Liu2026arxiv1985} instead uses the structure of the prediction problem. It splits action-conditioned prediction into state plausibility and action reachability, then enforces cycle consistency among proposed subgoals, inferred actions, and forward rollouts. Transitions that fail that check drive the update.

World-VLA-Loop~\cite{Liu2026arxiv6508} couples predictor repair to policy change, updating a video world model and a vision--language--action policy in the same cycle so each supplies training signal to the other. Its world model predicts future frames and binary rewards from the same diffusion latents, which attaches reward estimation to the generator rather than to a separate module. Coupling may accelerate adaptation, but it complicates attribution. An apparent model gain may come from a policy that has learned to avoid unsupported states, and an apparent policy gain may come from a newly introduced model bias.

These methods use prediction or verification feedback to update a world model. When model and policy updates are coupled, separating their contributions requires controlled comparisons.

\begin{takeaway}
The four improvement loops close different paths around the same grounded predictor. The data loop targets informative transitions, and the reward loop turns predicted outcomes into a score. The policy loop uses imagined experience, with support checks as one possible safeguard, and the model-self loop updates the predictor from predictive or verification feedback. These studies span static supervision, targeted feedback, and coupled updates; sustained gains must be established separately for each system. Influence and risk grow together, because a loop that can correct an error can also repeat it.
\end{takeaway}

\section{Embodiments and Domain Transfer}
\label{sec:domains}

The \Plevel{}--\Clevel{}--\Alevel{} ladder is shared across embodiments, but the variable that decides whether a claim holds is not. Manipulation is dominated by contact and object state. Driving is dominated by multi-agent futures and safety. Navigation is dominated by partial observability and persistent geometry. Locomotion is dominated by fast dynamics and stability.
The same reported number therefore carries different weight in each setting, and a method that meets the requirement in one domain need not meet it in another. This section takes a cross-cutting view, re-reading works from Sections~\ref{sec:background}--\ref{sec:actionable} by domain instead of by capability.



\subsection{Robotic Manipulation}
\label{sec:domain-manipulation}

Manipulation is decided by variables that pixels do not show. A predicted hand can reach the right object while the grasp pose, the friction, or the force trajectory is wrong.

The domain's answer has been to make contact an explicit predicted variable rather than a rendered consequence. ParticleFormer~\cite{Huang2025arxiv3126} predicts point-cloud dynamics for multi-material interaction, PIN-WM~\cite{Li2025arxiv6693} identifies 3D rigid-body dynamics from visual observation, and OmniVTA~\cite{Zheng2026arxiv9201} predicts contact evolution from tactile signals. Those signals carry contact forces and friction changes that vision alone does not reveal. FAWAM~\cite{He2026arxiv8555} extends the predicted variable to end-effector wrenches, so a contact error appears as a force residual rather than as a pixel difference. Long-horizon tasks add a second demand: the model must detect that a plan has failed and identify a state the policy can resume from. The recovery methods of Section~\ref{sec:actionable-verification} answer it in this domain. CycleVLA~\cite{Ma2026arxiv2295} backtracks to an earlier subtask, and See, Plan, Rewind~\cite{Dai2026arxiv9292} grounds an instruction into spatial milestones and returns to a recoverable state when progress stalls.

Evaluation in manipulation should compare rollouts that start from the same state and apply different actions. It should cover rigid and deformable objects, single- and bi-manual control, and both clean and perturbed execution. Long-horizon tasks should report subtask progress as well as final success, because that is what separates a planning failure from an execution failure or a failed recovery.

\subsection{Autonomous Driving}
\label{sec:domain-driving}

Driving world models must preserve road geometry, traffic rules, multi-agent intent, and the effect of ego control. The future is multimodal, so a model should not be penalized for failing to reproduce the single logged trajectory when several safe futures exist. It should instead place probability on feasible futures and preserve the risks that change the planner's choice.

Two routes appeared early and still organize the field. One keeps the predictive state compact and structured. MILE~\cite{Hu2022mile} learns latent driving dynamics with a geometric inductive bias, and OccWorld~\cite{Zheng2023arxiv6038} predicts in 3D occupancy space, a planning-compatible form. The other makes the generative view explicit. GAIA-1~\cite{Hu2023arxiv7080} treats driving as sequence modeling over video, text, and action tokens, and DriveDreamer~\cite{Wang2023arxiv9777} conditions diffusion-based generation on structured traffic constraints. Both routes were then turned toward the planner. Drive-WM~\cite{Wang2023arxiv7918} generates multi-view futures for distinct maneuvers so a planner can compare them. UniDrive-WM~\cite{Xiong2026arxiv4453} argues for one latent substrate shared by perception, prediction, and planning. DriveWorld-VLA~\cite{jia2026arxiv6521} uses the latent world state as the decision state, so control needs no pixel rollout. DriveVLA-W0~\cite{Li2026drivevlaw0} uses future-image prediction as dense self-supervision for end-to-end driving, and SteerVLA~\cite{Gao2026arxiv8440} treats a high-level model as a semantic world model that steers a low-level policy through long-tail maneuvers.

Closed-loop evaluation is needed to support driving actionability claims, and DriveArena~\cite{Yang2024arxiv0415} provides a generative simulation platform for it. The agreement of that simulator with road behavior is itself part of the claim. Open-loop displacement error against a single logged trajectory should not stand in for a closed-loop result. Studies should report route completion, collision, comfort, rule violations, intervention rate, and latency under one planner budget. Rare events need separate reporting, because average urban driving is a weak test of actionability.

\subsection{Navigation and Exploration}
\label{sec:domain-navigation}

Navigation makes memory part of plausibility. The agent must hold a stable belief over places and objects after they leave view, and the representation must support loop closure and distinguish perceptual aliasing. Geometry may be metric~\cite{yu2025rgbonlygaussiansplattingslam,
cheng2025outdoormonocularslamglobal}, topological, semantic, or implicit; what matters is that the belief survives.

The lineage runs from imagining unseen observations to rolling them out under candidate actions. Pathdreamer~\cite{Koh2021pathdreamer} generates plausible panoramic RGB, depth, and semantic observations for viewpoints the agent has not visited. The navigation model VISTA~\cite{Huang2025arxiv7868} makes that imagination instruction-conditioned, and VISTAv2~\cite{Huang2025arxiv0041} rolls out egocentric futures under candidate actions and projects them into an online value map. Navigation World Models~\cite{Bar2025nwm} state the same problem as controllable video generation, and SparseVideoNav~\cite{Zhang2026arxiv5827} replaces dense long-horizon rollout with sparse future generation for faster deployment. EgoWM~\cite{Bagchi2026arxiv5284} adapts internet-scale video diffusion into an action-conditioned egocentric predictor. X-MOBILITY~\cite{Liu2024arxiv7491} instead separates world modeling from the action policy so that off-policy data trains the dynamics and on-policy data trains control.

This lineage shows that the useful part of a navigation world model is the hidden structure it exposes: future visibility, traversability, and goal progress. A benchmark should therefore separate map accuracy from navigation utility, since one model may render a complete map while another retains only the structure needed to reach the goal. Exploration is where that difference shows, because it scores the belief by what the agent can reach rather than by what the model can draw.

\subsection{Locomotion and Control}
\label{sec:domain-locomotion}

Locomotion is bounded by the control clock. The predictor must return a state fast enough to sit inside the control loop, which is why the domain uses compact states more often than decoded video. Plausibility here concerns short-horizon dynamics, terrain contact, and stability, and controllability concerns the effects of torque, velocity commands, or latent skills. Actionability is read from return, fall rate, constraint violation, and adaptation to terrain or morphology.

The compact-state models this domain inherits were developed for continuous-control benchmarks and games rather than for legged hardware. PlaNet~\cite{Hafner2018arxiv4551} plans online in a learned latent state space, and Dreamer~\cite{Hafner2019arxiv1603} learns policies from trajectories imagined in that space. DayDreamer~\cite{Wu2022arxiv4176} then carried the same latent imagination onto physical robots. DreamerV3~\cite{Hafner2023arxiv4104} solves more than 150 diverse tasks with a single configuration, and TD-MPC2~\cite{Hansen2023arxiv6828} combines a learned latent model with local trajectory optimization across task domains, embodiments, and action spaces. These systems use compact predictive states for planning or policy learning; the cost of visual rollout remains a separate deployment consideration.

A slow visual generator may serve high-level planning, but it cannot replace a high-rate dynamics model inside the control loop. Hierarchical designs may run different models at different rates, provided their states and uncertainties stay aligned, and a report should distinguish the rate of sensing, prediction, planning, and low-level control.

\subsection{What Transfers Across Embodiments}
\label{sec:domain-transfer}

The four domains put different variables in front of the same ladder. Manipulation predicts contact itself, in ParticleFormer~\cite{Huang2025arxiv3126}, PIN-WM~\cite{Li2025arxiv6693}, OmniVTA~\cite{Zheng2026arxiv9201}, and FAWAM~\cite{He2026arxiv8555}. Driving predicts a planner-facing scene state, in OccWorld~\cite{Zheng2023arxiv6038}, Drive-WM~\cite{Wang2023arxiv7918}, and DriveWorld-VLA~\cite{jia2026arxiv6521}. Navigation predicts a belief that outlives the view, in Pathdreamer~\cite{Koh2021pathdreamer}, VISTAv2~\cite{Huang2025arxiv0041}, and Navigation World Models~\cite{Bar2025nwm}. Locomotion and control use compact predictive states, in PlaNet~\cite{Hafner2018arxiv4551}, Dreamer~\cite{Hafner2019arxiv1603}, and TD-MPC2~\cite{Hansen2023arxiv6828}.

Whether any of those predictions carries to another embodiment is a separate question, and it can be asked of four things: representation, dynamics, action semantics, or decision logic. Object permanence and free-space geometry transfer widely. Contact dynamics transfer only after material, tool, and actuator differences are accounted for. A language goal may be shared across embodiments while its action realization is not.

The interfaces built for transfer make the distinction concrete. Object-level motion interfaces such as 3DFlowAction~\cite{Zhi2025arxiv6199} and the object-centric 3D motion field~\cite{Yin2025objectcentric} state the command as the future motion of the manipulated object. The command is then defined independently of the robot that executes it. Latent-action models take the complementary route. Genie~\cite{Bruce2024arxiv5391} recovers a control variable from unlabeled video. AdaWorld~\cite{Gao2025arxiv8938} conditions a pre-trained world model on such variables, so the model adapts to environments with heterogeneous action spaces. LDA-1B~\cite{Lyu2026arxiv2215} scales latent action learning by ingesting embodied data from many sources. Cosmos~\cite{NVIDIA2025arxiv3575} supplies pre-trained world foundation models that are post-trained for a target setting. A transferable interface and measured transfer remain different claims, and only the second requires matched robots, objects, and tasks.

Three protocols test transfer. Hold the task fixed and change the embodiment. Hold the embodiment fixed and change objects, scenes, or dynamics. Change both, and measure how much grounded adaptation is needed. Across all three, the report should state which modules are frozen, which are adapted, and how many real transitions the adaptation consumed. 

A bottleneck is that these domains emphasize different outcome measures. A manipulation result is read in subtask progress, a driving result in closed-loop route outcomes, a navigation result in goal progress, and a locomotion result in return and fall rate. Adaptation cost can complement these domain-specific outcomes in cross-domain transfer reports. Section~\ref{sec:data-evaluation} takes up the data and evaluation protocols that would make such reports comparable, and Section~\ref{sec:open-problems} states what remains unresolved.

\begin{takeaway}
The capability ladder transfers across domains, but the test variables and operating conditions must match the embodiment. The domains differ in which variable decides the claim: contact and object state in manipulation, multi-agent futures in driving, persistent belief in navigation, and body dynamics at control rate in locomotion. The transfer literature has produced interfaces that are defined independently of the robot, in object-level motion commands and in latent actions recovered from video. Comparing these interfaces across embodiments calls for consistent reporting of adaptation cost.
\end{takeaway}

\section{Data, Benchmarks, and Evaluation}
\label{sec:data-evaluation}

Evaluation should follow the capability claim. A rollout that preserves task-relevant state supports a plausibility claim. An intervention test supports a controllability claim. A better decision or learning update supports an actionability claim. No single metric proves all three.

\subsection{Data Requirements}

A world-model trajectory links an initial state, observations, actions, time, and resulting state. Useful records may also include language goals, camera calibration, depth, proprioception, force, contact, audio, reward, failure onset, and recovery outcome. The right fields depend on the grounding claim, but timing and action semantics are always part of the data.

RoboNet~\cite{dasari2019robonet} introduced a multi-robot dataset for action-conditioned video prediction and visual planning. Open X-Embodiment~\cite{oneill2024openx} aggregates trajectories from many robot platforms. BridgeData V2~\cite{walke2023bridgedatav2} and DROID~\cite{khazatsky2024droid} provide broad real-world manipulation data with aligned observations and actions. These corpora support action grounding, but cross-embodiment use requires control rate, coordinate frames, calibration, and the original action definition. A normalized action vector alone does not define a shared intervention.

Structured simulators enable counterfactual data because the same state can be replayed under different actions. RLBench~\cite{james2020rlbench}, CALVIN~\cite{mees2022calvin}, and LIBERO~\cite{liu2023libero} provide reproducible manipulation tasks and executable outcome checks. RH20T~\cite{fang2023rh20t} adds visual, force, and audio signals for contact-rich skills. These resources cover complementary variables: task diversity, long-horizon composition, generalization, and multimodal contact. A study should select data because it identifies the target transition, not because it is the largest available corpus.

Success-only demonstrations leave verification and recovery weakly supervised. A closed-loop dataset should record the failure onset, likely cause, recoverable state, corrective action, and final outcome. Paired nominal and perturbed runs are especially useful because they isolate the event that invalidated the original prediction. Synthetic failures can extend coverage, but they should be checked for physical and task validity before policy training.

\begin{table*}[t]
\centering
\caption{
Concrete evaluation protocols for the three capability levels. Each level is paired with representative datasets, commonly used metrics, and the key control needed to support the corresponding claim. Datasets are illustrative rather than mandatory, and higher-level claims should retain relevant lower-level tests.
}
\label{tab:evaluation-ladder}
\scriptsize
\setlength{\tabcolsep}{3.5pt}
\renewcommand{\arraystretch}{1.18}

\begin{tabularx}{\textwidth}{
    >{\raggedright\arraybackslash\bfseries}p{0.100\textwidth}
    >{\raggedright\arraybackslash}p{0.225\textwidth}
    >{\raggedright\arraybackslash}p{0.235\textwidth}
    >{\raggedright\arraybackslash}X
    >{\raggedright\arraybackslash}p{0.17\textwidth}}
\toprule
Claim
& Concrete test
& Typical datasets / benchmarks
& Main metrics
& Key control \\
\midrule

Plausible
&
Roll out future observations or states on unseen episodes and evaluate whether appearance, geometry, objects, and physical relations remain consistent.
&
Physion \cite{Bear2021arxiv8261};
IntPhys \cite{Riochet2018arxiv7616};
PHYRE \cite{Bakhtin2019arxiv5656};
RoboNet \cite{dasari2019robonet};
BridgeData V2 \cite{walke2023bridgedatav2}.
&
SSIM~\cite{wang2004ssim}, LPIPS~\cite{zhang2018lpips}, and FVD~\cite{unterthiner2018fvd}; depth/occupancy error; object and contact accuracy; drift versus horizon.
&
Persistence and short-context baselines; memory ablation; unseen scenes or dynamics.
\\
\midrule

Controllable
&
Start from the same state, vary one action, force, timing, trajectory, or goal, and compare the predicted intervention outcome with execution.
&
RLBench \cite{james2020rlbench};
CALVIN \cite{mees2022calvin};
LIBERO \cite{liu2023libero};
ACT-Bench \cite{Arai2024arxiv5337};
MiraBench \cite{Yang2026arxiv9360}.
&
Action adherence; state-change error; trajectory ADE/FDE; counterfactual ranking; unchanged-region consistency.
&
Masked, shuffled, wrong, and no-action inputs; shared initial states.
\\
\midrule

Actionable
&
Use the model for planning, policy learning, evaluation, verification, or recovery, and measure the resulting closed-loop behavior.
&
RLBench \cite{james2020rlbench};
CALVIN \cite{mees2022calvin};
LIBERO \cite{liu2023libero};
DriveArena \cite{Yang2024arxiv0415};
WorldGym \cite{Quevedo2025arxiv0613};
WorldArena \cite{Shang2026arxiv8971}.
&
Task success or return; planning regret; collision or safety cost; latency and model calls; detection and recovery success.
&
Same policy without the model; oracle model; equal data and compute budgets.
\\

\bottomrule
\end{tabularx}
\end{table*}

\subsection{A Capability-Aligned Evaluation Protocol}

\paragraph{Plausibility and stability.}
Plausibility is evaluated under fixed observation or action context. Image metrics such as SSIM~\cite{wang2004ssim} and LPIPS~\cite{zhang2018lpips}, and video metrics such as FVD~\cite{unterthiner2018fvd}, measure appearance. They do not measure causal correctness. Grounded measures add object identity, 3D position, depth, occupancy, topology, contact state, momentum, and constraint violations. Long-horizon tests report error as a function of rollout time, not only one average. They also report memory retention after occlusion and the frequency of irreversible drift.

\paragraph{Controllability and intervention consistency.}
A controllable model is tested with paired or branched interventions. Starting from the same state, the evaluator changes the action, action timing, route, force, or trajectory constraint. The model should preserve what did not change and alter the variables caused by the intervention. Metrics include action adherence, treatment-effect error, counterfactual ranking, contact consistency, and invariance of unrelated scene content. Action-shuffled and action-masked controls show whether the model actually uses the command.

\paragraph{Actionability and decision utility.}
Actionability is evaluated inside its decision path. Planning reports task success, regret, model calls, replanning rate, and latency. Reinforcement learning reports realized return, constraint cost, real interaction count, and the ratio of imagined to grounded data. Policy evaluation reports rank correlation and calibration against real outcomes. Verification reports precision--recall, false alarms per episode, detection delay, and risk--coverage. Correction and recovery report conditional recovery success, retained progress, extra actions, recovery time, and safety violations.

Under the proposed protocol, each result should include a matched baseline without world-model input. Additional controls may use a reactive monitor, a generic replanning trigger, or an oracle predictor. These comparisons identify whether the gain comes from predictive structure, extra computation, or a stronger policy backbone.



\subsection{Benchmark Design Principles}

First, benchmark state should be executable. When possible, a predicted manipulation, route, or body state is checked in a grounded simulator or physical system. Second, the benchmark should expose branching actions from shared initial states. Third, visual quality and functional utility are reported separately. Fourth, perturbations are parameterized and repeatable. Fifth, the benchmark records model calls and wall-clock time because a correct future that arrives after the control deadline is not actionable.

Uncertainty deserves a protocol rather than one scalar. Studies should report calibration over horizon, action-dependent confidence, selective prediction, and risk under a fixed coverage level. For generative models, uncertainty from diverse samples should be separated from uncertainty caused by weak dynamics. Diversity is useful only when the set of samples covers feasible outcomes and preserves their relative risks.

Finally, comparisons should use the same observation history, action proposal set, and downstream decision budget. If a method uses extra cameras, privileged simulator state, or more model calls, these resources are part of the method and must appear in the main table.

\begin{takeaway}
Data and evaluation must match the claimed capability and grounding. Plausibility needs held-out rollouts, state-based errors, drift curves, memory tests, and calibrated uncertainty. Controllability needs shared-start intervention branches and masked, shuffled, or wrong-action controls. Actionability needs matched controllers, equal decision budgets, realized utility, latency, safety, and failure costs. Datasets should preserve timing, action semantics, unsuccessful episodes, and clean transfer splits. No single visual metric spans the ladder; each higher claim requires stronger controls and a grounded outcome.
\end{takeaway}

\section{Open Problems}
\label{sec:open-problems}

The main open problems sit at the boundaries between capability levels. A model may cross one boundary in a narrow setting and still fail under a new embodiment, time scale, or interaction regime. We group the gaps by the proof that is still missing, and each subsection names the test that would supply it.

\subsection{Persistent State and Drift}
\label{sec:open-state}

Current models generate convincing short futures, but identity, geometry, and dynamics drift over long rollouts. More context alone does not guarantee consistency. Long-context architectures such as state-space video world models~\cite{Po2025arxiv0171} and distillation schemes such as LongDWM~\cite{Wang2025arxiv1546} extend rollout length, but longer rollouts alone do not establish state persistence. Memory mechanisms address the right variable: persistent embodied world models~\cite{Zhou2025arxiv5495} and long-term spatial memory~\cite{Wu2025arxiv5284} store content so that it can be recovered. An open challenge is a state that can be corrected without rewriting consistent history, together with uncertainty that grows while a state stays unresolved.

The test is a hide-and-reveal protocol. Occlude an object, change other parts of the scene, and reveal it later. The model should track identity and pose using observations and action history, and retain uncertainty when those inputs do not determine the hidden state. The same protocol applies to physical properties such as mass, friction, and articulation, which PIN-WM~\cite{Li2025arxiv6693} estimates but which few predictors are asked to retain. A long-horizon score should show when an error starts and whether it can be repaired.

\subsection{Causal Action Effects}
\label{sec:open-causal}

Large video models learn strong temporal priors, and a correlation between an action and a future is not an intervention model. Confounding enters through demonstrations, because a skilled agent chooses actions that already suit the scene and the goal. A predictor can then reproduce the common outcome while ignoring the command, which is why ACT-Bench~\cite{Arai2024arxiv5337} and WorldSimProbe~\cite{Co2026arxiv9298} measure action realization separately from scene quality.

Progress needs data with branch points, action perturbations, and failed actions, and tests that hold the initial state fixed while one action variable changes. Cross-embodiment transfer sharpens the problem, because one semantic action has different kinematics and timing on different robots. Latent-action models such as LAPA~\cite{Ye2024arxiv1758} and AdaWorld~\cite{Gao2025arxiv8938} give the interface, and scene-level motion commands such as 3DFlowAction~\cite{Zhi2025arxiv6199} give an embodiment-independent alternative. Transfer evaluation should test whether intent and embodiment-specific realization are preserved together.

\subsection{Decision Utility Under Budget}
\label{sec:open-utility}

A model can respond correctly to actions and still be unusable for planning. It may be too slow, uncertain in the region that matters, or wrong about the small differences that reverse an action ranking. Latency is the constraint the field has begun to design around. AHA-WAM~\cite{Cai2026arxiv9811} decouples prediction from execution rhythm, and ForeWAM~\cite{Huang2026arxiv1605} supplies predictive context without decoding video. Fast-WAM~\cite{Yuan2026arxiv6666} asks whether test-time imagination is needed at all. FutureRTC~\cite{Jiang2026arxiv4008} predicts the state that will hold when a chunk finally executes.

A remaining challenge is decision-aware compression that preserves reward, feasibility, or safety information while retaining enough grounded state to detect model exploitation. The benchmark that would settle it is regret under a fixed compute budget, comparing the action selected against the action a grounded outcome would justify. Varying proposal quality, horizon, and model-call budget would then show when decoded video, latent state, object state, or a 3D scene is the right interface.

\subsection{Joint Grounding}
\label{sec:open-grounding}

Most systems are strong on one or two grounding axes. Geometry-first predictors such as OccWorld~\cite{Zheng2023arxiv6038} and GaussianWorld~\cite{Zuo2024arxiv0373} may miss contact. Physics-oriented models such as PhyDNet~\cite{Guen2020arxiv1460} and ParticleFormer~\cite{Huang2025arxiv3126} carry dynamics without semantic memory. Action-conditioned generators may follow a command without preserving 3D structure. Systems that combine axes exist, including 3D-VLA~\cite{Zhen2024arxiv9631} and OmniVTA~\cite{Zheng2026arxiv9201}, but combining axes does not by itself establish consistency between their predictions.
The next step is an interface on which specialized representations agree about state, uncertainty, and time, rather than one universal representation.

Evaluation should therefore include cases where the axes conflict. A geometrically reachable grasp may be physically unstable. A physically possible motion may violate the instruction. An action that is valid in one coordinate frame may be wrong after a camera or embodiment change. Such cases reveal whether grounding modules exchange causal state or only concatenate features.

\subsection{Uncertainty and Stable Loops}
\label{sec:open-uncertainty}

Uncertainty is usually measured on held-out data from the training distribution. An actionable model faces policy shift, because the controller visits states precisely where the model predicted they would be useful. That feedback channel can make ordinary calibration fail. Calibrated confidence inside the generated frame~\cite{Mei2025arxiv5927} and conformal treatments of failure detection~\cite{Ward2026arxiv6987,Ho2026arxiv6182} offer partial answers; confidence should also be re-validated after policy updates.

A world model needs action-dependent and horizon-dependent confidence, and defined behavior when confidence is low: shorter rollout, conservative value, a request for data, a fallback policy, or a safe stop. The same requirement compounds across the four loops of Section~\ref{sec:loops}. Synthetic data can teach a model its own artifacts~\cite{11543513}, a critic can reward them~\cite{yao2026motiongrpo}, and a policy can then seek them. RLVR-World~\cite{Wu2025arxiv3934} and World Action Verifier~\cite{Liu2026arxiv1985} introduce external or structural checks, and World-VLA-Loop~\cite{Liu2026arxiv6508} and VLAW~\cite{Guo2026arxiv2063} couple the updates without separating the credit. A loop-level benchmark should run several data, model, and policy cycles, measure improvement, regression, diversity, and interaction cost after each, and hold out a set of dynamics and failures the generator cannot inspect.

\subsection{Evaluation, Access, and Reproducibility}
\label{sec:open-evaluation}

Manipulation dominates recent world-action work, while navigation, locomotion, and driving keep separate benchmarks and vocabulary. A shared layer could measure state stability, intervention consistency, decision regret, calibration, and closed-loop utility, with domain layers adding contact force, collision, comfort, fall rate, map consistency, or route completion. WorldArena~\cite{Shang2026arxiv8971} and RoboWM-Bench~\cite{Jiang2026arxiv9092} move in this direction for manipulation, and World-in-World~\cite{Zhang2025arxiv8135} shows what a standardized closed-loop interface makes comparable. Real-system evaluation should also report enough trials and uncertainty estimates to assess variability and rare failures.

Timing is a second axis that resists shared reporting. A high-level generator may predict seconds ahead while contact control needs updates within milliseconds, so a hierarchical system needs an explicit contract between slow semantic prediction, mid-level action chunks, and fast feedback. That contract covers state timestamps, committed action prefixes, uncertainty, and cancellation rules. Benchmarks should inject sensing delay and variable inference time, then report deadline misses and stale-action execution beside average latency, as the asynchronous-deployment study of Section~\ref{sec:actionable-verification}~\cite{Team2026arxiv1880} does for one platform.

Access is the last barrier. Many recent systems depend on large video backbones such as Cosmos~\cite{NVIDIA2025arxiv3575}, on private robot data, or on expensive real trials. Reproducibility needs more than a released checkpoint. A study should publish its action interface, data filters, model-call budget, inference latency, evaluation seeds, and failure cases. Smaller diagnostic models and frozen benchmark traces can test a claim before anyone pays for full-system training.

\begin{takeaway}
Progress depends on closing the two gaps in the ladder and keeping the closed loops stable. Long-horizon plausibility needs a correctable persistent state and uncertainty that grows while a state is unresolved. Controllability needs branched data and an action representation that survives a change of embodiment. Actionability needs decision-aware compression and evaluation under a real-time budget. Geometry, physics, and action modules must agree when their predictions conflict, and repeated loop updates need an independent audit to keep a shared error from circulating.
\end{takeaway}

\section{Conclusion}

For embodied intelligence, the value of a world model lies in what its predictions enable an agent to do. Guided by this principle, this survey organizes the literature around a decision-centered capability ladder: a \Plevel{} model preserves task-relevant temporal, geometric, or physical structure; a \Clevel{} model predicts how interventions change that structure; and an \Alevel{} model uses prediction to produce measurable downstream gains. We complement this ladder with a $3\times4$ conceptual matrix linking geometry, physics, and action grounding to data, reward, policy, and model-self improvement loops. Together, these views organize representative methods across manipulation, navigation, locomotion, autonomous driving, and general embodied learning. Our review of datasets, benchmarks, and evaluation protocols highlights the need to assess state consistency, intervention fidelity, and downstream utility separately. The framework supports qualitative comparison of capability claims, while heterogeneous evaluation settings limit direct comparison of reported results. Open challenges include long-horizon consistency, uncertainty calibration, causal intervention testing, latency, verification and recovery, and cross-embodiment transfer. Progress toward actionability should be evaluated by whether grounded predictions improve closed-loop behavior under explicit data, compute, and latency budgets.

\bibliographystyle{plainnat}
\bibliography{references}

\end{document}